\documentclass[10pt,twocolumn,letterpaper]{article}

\usepackage{cvpr}              

\usepackage[accsupp]{axessibility}
\usepackage{graphicx}
\usepackage{float}
\usepackage{amsmath}
\usepackage{amssymb}
\usepackage{booktabs}
\usepackage{stmaryrd}
\usepackage{xcolor}
\usepackage{amsthm}
\usepackage{multicol}
\usepackage[symbol]{footmisc}

\usepackage[pagebackref,breaklinks,colorlinks]{hyperref}

\usepackage[capitalize]{cleveref}
\crefname{section}{Sec.}{Secs.}
\Crefname{section}{Section}{Sections}
\Crefname{table}{Table}{Tables}
\crefname{table}{Tab.}{Tabs.}

\newtheorem{Def sec}{Definition}[section]   
\newtheorem{Definition}{Definition}[section]
\newtheorem{Lemma}{Lemma}[section]

\usepackage{blindtext}
\usepackage{multirow}

\def\cvprPaperID{807} 
\def\confName{CVPR}
\def\confYear{2022}

\usepackage{wrapfig}
\usepackage{ifthen}
\newboolean{showNotes}
\setboolean{showNotes}{false}

\usepackage{xspace}

\newcommand{\ShortName}{ConDor\xspace}
\newcommand{\Rahul}[1]{\textbf{\textcolor{red}{Rahul: #1}}}
\newcommand{\Srinath}[1]{\textbf{\textcolor{blue}{Srinath: #1}}}
\newcommand{\Adrien}[1]{\textbf{\textcolor{orange}{Adrien: #1}}}
\newcommand{\Radhika}[1]{\textbf{\textcolor{brown}{Radhika: #1}}}
\newcommand{\Jivitesh}[1]{\textbf{\textcolor{olive}{Jivitesh: #1}}}
\newcommand{\Leo}[1]{\textbf{\textcolor{purple}{Leo: #1}}}

\crefname{section}{Section}{Section}
\crefname{equation}{Equation}{Equation}
\crefname{figure}{Figure}{Figure}
\crefname{table}{Table}{Table}

\newcommand{\parahead}[1]{\noindent\textbf{#1}:\ }

\newenvironment{packed_itemize}
{\begin{itemize}
    \setlength{\itemsep}{1pt}
    \setlength{\parskip}{0pt}
    \setlength{\parsep}{0pt}
}{\end{itemize}}

\newcommand{\filluptopage}[1]{%
  \clearpage
  \loop\ifnum\value{page}<#1\relax
    \null\clearpage
  \repeat
  \loop\ifnum\value{page}=#1\relax
    \null\clearpage
  \repeat
}

\makeatletter
\def\blfootnote{\xdef\@thefnmark{}\@footnotetext}
\makeatother

\begin{document}

\title{\ShortName: Self-Supervised Canonicalization of 3D Pose for Partial Shapes}

\author{
Rahul Sajnani$^{1}$\qquad Adrien Poulenard$^{2}$\qquad Jivitesh Jain$^{1}$ \qquad Radhika Dua$^{3}$\\ \qquad Leonidas J. Guibas$^{2}$ \qquad Srinath Sridhar$^{4}$ \\ \vspace{1mm}
\text{\normalsize $^1$RRC, IIIT-Hyderabad\qquad $^2$Stanford University\qquad $^3$KAIST\qquad  $^4$Brown University}\\
\href{https://ivl.cs.brown.edu/ConDor/}{ivl.cs.brown.edu/ConDor}
}

\twocolumn[{%
\renewcommand\twocolumn[1][]{#1}%
\maketitle
\begin{center}
    \vspace{-1cm}
    \centering
    \captionsetup{type=figure}
    \includegraphics[width=\textwidth]{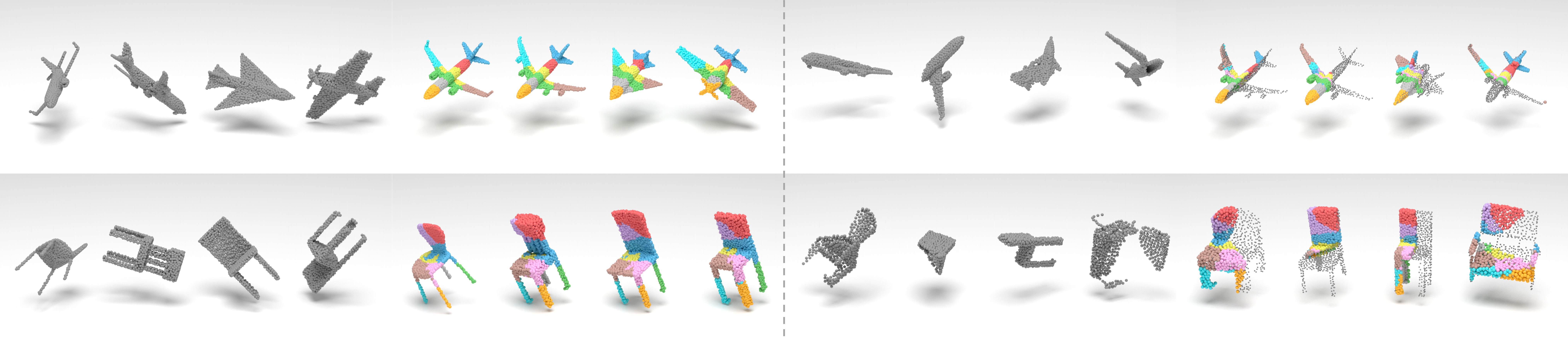}
    \vspace{-0.25in}
    \captionof{figure}{
    \textbf{\ShortName} is a self-supervised method that learns to \underline{\textbf{C}}an\underline{\textbf{on}}icalize the 3\underline{\textbf{D or}}ientation and position (3D pose) for full and partial shapes.
    (\emph{\textbf{left}})~Our method takes un-canonicalized 3D point clouds (gray) from different categories as input and produces consistently canonicalized outputs (colored).
    (\emph{\textbf{right}})~Our method can also operate on partial point clouds
    (missing part of shape shown only for visualization).
    In addition, \ShortName can also learn consistent co-segmentation of shapes without supervision, visualized as colored parts.
    \vspace{-0.2cm}
    }
    \label{fig:teaser}
\end{center}%
}]

\begin{abstract}
Progress in 3D object understanding has relied on manually ``canonicalized'' shape datasets that contain instances with consistent position and orientation (3D pose).
This has made it hard to generalize these methods to in-the-wild shapes, \eg,~from internet model collections or depth sensors.
\textbf{\ShortName} is a self-supervised method that learns to \underline{\textbf{C}}an\underline{\textbf{on}}icalize the 3\underline{\textbf{D or}}ientation and position for full and partial 3D point clouds.
We build on top of Tensor Field Networks (TFNs), a class of permutation- and rotation-equivariant, and translation-invariant 3D networks.
During inference, our method takes an unseen full or partial 3D point cloud at an arbitrary pose and outputs an equivariant canonical pose.
During training, this network uses self-supervision losses to learn the canonical pose from an un-canonicalized collection of full and partial 3D point clouds.
\ShortName can also learn to consistently co-segment object parts without any supervision.
Extensive quantitative results on four new metrics show that our approach outperforms existing methods while enabling new applications such as operation on depth images and annotation transfer.
\end{abstract}

\vspace{-0.2in}
\section{Introduction}
\label{sec:intro}
%
%
Humans have the ability to recognize 3D objects in a wide variety of positions and orientations (poses)~\cite{shepard1971mental}, even if objects are occluded.
We also seem to prefer certain \emph{canonical views}~\cite{cutzu1994canonical}, with evidence indicating that an object in a new pose is \emph{mentally rotated} to a canonical pose~\cite{tarr1989mental} to aid recognition.
%
Inspired by this, we aim to build scene understanding methods that reason about objects in different poses by learning to map them to a canonical pose without explicit supervision.

Given a 3D object shape, the goal of \textbf{instance-level canonicalization} is to find an \emph{equivariant frame} of reference that is consistent relative to the geometry of the shape under different 3D poses.
This problem can be solved if we have shape correspondences and a way to find a distinctive equivariant frame (\eg,~PCA).
However, it becomes significantly harder if we want to operate on different 3D poses of different object instances
that lack correspondences.
This \textbf{category-level canonicalization} problem has received much less attention despite tremendous interest in category-level 3D object understanding~\cite{wu20153d, choy20163d, park2019deepsdf, mescheder2019occupancy, groueix2018papier, deprelle2019learning, mildenhall2020nerf}.
Most methods rely on data augmentation~\cite{liu2020fg}, or manually annotated datasets~\cite{chang2015shapenet, wu20153d} containing instances that are consistently positioned and oriented within each category~\cite{wang2019normalized, tatarchenko2019single, sridhar2019multiview}.
This has prevented broader application of these methods to un-canonicalized data sources, such as online model collections~\cite{UnityAss22:online}.
The problem is further exacerbated by the difficulty of canonicalizing partial shape observations (\eg,~from depth maps~\cite{reizenstein2021common}), or symmetric objects that require an understanding of inter-instance part relationships.
Recent work addresses these limitations using weakly-supervised~\cite{sajnani2021draco, gu2020weaklysupervised} or self-supervised learning~\cite{novotny2019c3dpo, sun2020canonical, spezialetti2020learning, rotpredictor}, but cannot handle partial 3D shapes, or is limited to canonicalizing only orientation.

We introduce \textbf{\ShortName}, a method for self-supervised category-level \underline{\textbf{C}}an\textbf{\underline{on}}icalization of the 3\underline{\textbf{D}} p\underline{\textbf{o}}se of pa\underline{\textbf{r}}tial shapes.
It consists of a neural network that is trained on an un-canonicalized collection of 3D point clouds with inconsistent 3D poses.
During inference, our method takes a full or partial 3D point cloud of an object at an arbitrary pose, and outputs a canonical rotation frame and translation vector.
To enable operation on instances from different categories, we build upon Tensor Field Networks (TFNs)~\cite{thomas2018tensor}, a 3D point cloud architecture that is equivariant to 3D rotation and point permutation, and invariant to translation.
To handle partial shapes, we use
a two-branch (Siamese) network with training data that simulates partiality through shape slicing or camera projection.
We introduce several losses to help our method learn to canonicalize 3D pose via self-supervision.
A surprising feature of our method is the (optional) ability to learn consistent part co-segmentation~\cite{chen2019bae} across instances without any supervision (see \cref{fig:teaser}).


Given only the recent interest, \textbf{standardized metrics} for evaluation of canonicalization methods have not yet emerged.
We therefore propose four new metrics that are designed to evaluate the consistency of instance- and category-level canonicalization, as well as consistency with manually pre-canonicalized datasets.
We extensively evaluate the performance of our method using these metrics by comparing with baselines and other methods~\cite{sun2020canonical, spezialetti2020learning}.
Quantitative and qualitative results on common shape categories show that we outperform existing methods and produce consistent pose canonicalizations for both full and partial 3D point clouds.
We also demonstrate previously difficult \textbf{applications} enabled by our method such as operation on partial point clouds from depth maps, keypoint annotation transfer, and expanding the size of existing datasets.
To sum up, our contributions include:

%
\begin{packed_itemize}
    \item A self-supervised method to canonicalize the 3D pose of full point clouds from a variety of object categories.
    \item A method that can also handle \textbf{partial} 3D point clouds.
    \item New metrics to evaluate canonicalization methods, extensive experiments, and new applications.
\end{packed_itemize}
\section {Related Work}
\label{sec:relwork}
Canonical object representations in human perception have been extensively studied as mental rotation~\cite{shepard1971mental, tarr1989mental}, shape constancy and equivalence~\cite{palmer1999vision}, and canonical views~\cite{cutzu1994canonical}.
We review related work that studies or uses canonicalization for machine perception of 3D scenes.

\parahead{3D Scene Understanding}
%
Invariance and equivariance to 3D pose in tasks such as shape classification, reconstruction and registration was initially achieved using hand-crafted features~\cite{johnson1997spin, rusu2009fast, salti2014shot}.
With machine learning, these features were replaced with learned features~\cite{wang2015voting, deepgmr, wang2019deep}, but 3D data introduces challenges in learning invariant features~\cite{qi2017pointnet}.
Data augmentation by sampling the space of 3D poses for each object is one way to address this problem~\cite{liu2020fg} but results in longer training and larger networks.
Category-level object reconstruction methods have gained significant attention with representations ranging from voxel grids~\cite{choy20163d,mescheder2019occupancy}, implicit surfaces~\cite{park2019deepsdf}, parametric surfaces~\cite{groueix2018papier, lei2020pix2surf}, point clouds~\cite{yang2019pointflow}, and depth images~\cite{zhang2018learning}.
Almost all of these methods rely on manually pre-canonicalized datasets like ShapeNet~\cite{shapenet2015} and ModelNet40~\cite{wu20153d} to learn inductive biases for effective learning~\cite{tatarchenko2019single}.
Neural networks have also been successfully used for supervised~\cite{mo2019partnet} and unsupervised~\cite{chen2019bae} segmentation of object parts.

\parahead{3D Neural Networks}
Numerous neural networks have been proposed for processing 3D data represented as
voxels~\cite{maturana2015voxnet, qi2016volumetric, wu20153d}, multiple views~\cite{su2015multi}, point clouds\cite{qi2017pointnet, qi2017pointnetpp, wang2018dynamic} or meshes~\cite{hanocka2019meshcnn, yang2021continuous}.
For 3D point clouds, PointNet and related methods achieve point permutation equivariance and translation equivariance, but not rotation equivariance.
Spherical CNNs~\cite{cohen2018spherical} and Tensor Field Networks (TFNs)~\cite{thomas2018tensor, poulenard2021functional} address this limitation.
We use 3D point clouds as our shape representation and TFNs to achieve equivariance to permutation, translation, and rotation.

\parahead{Supervised Canonicalization}
Supervised canonicalization of shapes enables applications such as instance-level camera pose estimation~\cite{shotton2013scene} or human pose estimation~\cite{shotton2011real, keskin2013real}.
It can also be useful for \emph{category-level} reasoning, for example 6~DoF pose estimation~\cite{wang2019normalized, wang20206}.
However, these methods are limited to learning from data with ground truth canonicalization making it hard to generalize to real data.
%


%
Our method is most related to recent work on weakly supervised~\cite{sajnani2021draco, gu2020weaklysupervised}, or self-supervised learning of canonicalization of semantic keypoints ~\cite{novotny2019c3dpo} and point clouds~\cite{sun2020canonical, spezialetti2020learning}.
Unlike these methods, we can canonicalize both orientation and translation for partial shapes.

\section{Background}
\label{sec:background}

\parahead{3D Pose Canonicalization\label{sec:pose_canon}}
The 3D pose of an object refers to its 3D position and orientation in space specified by an intrinsic object-centric reference frame (defined by an origin and orthonormal rotation).
Having a consistent intrinsic frame across different shapes is critical in many problems~\cite{esteves2018learning, chen2019clusternet, zhang2019rotation, poulenard2021functional}.
We denote such a consistent intrinsic frame as a \textbf{canonical frame}.
This frame transforms together with the object, \ie,~it is equivariant.
The object pose is constant relative to the canonical frame -- we call this our \textbf{canonical pose}.
%
\begin{figure}[h!]
\centering
  \includegraphics[width=\columnwidth]{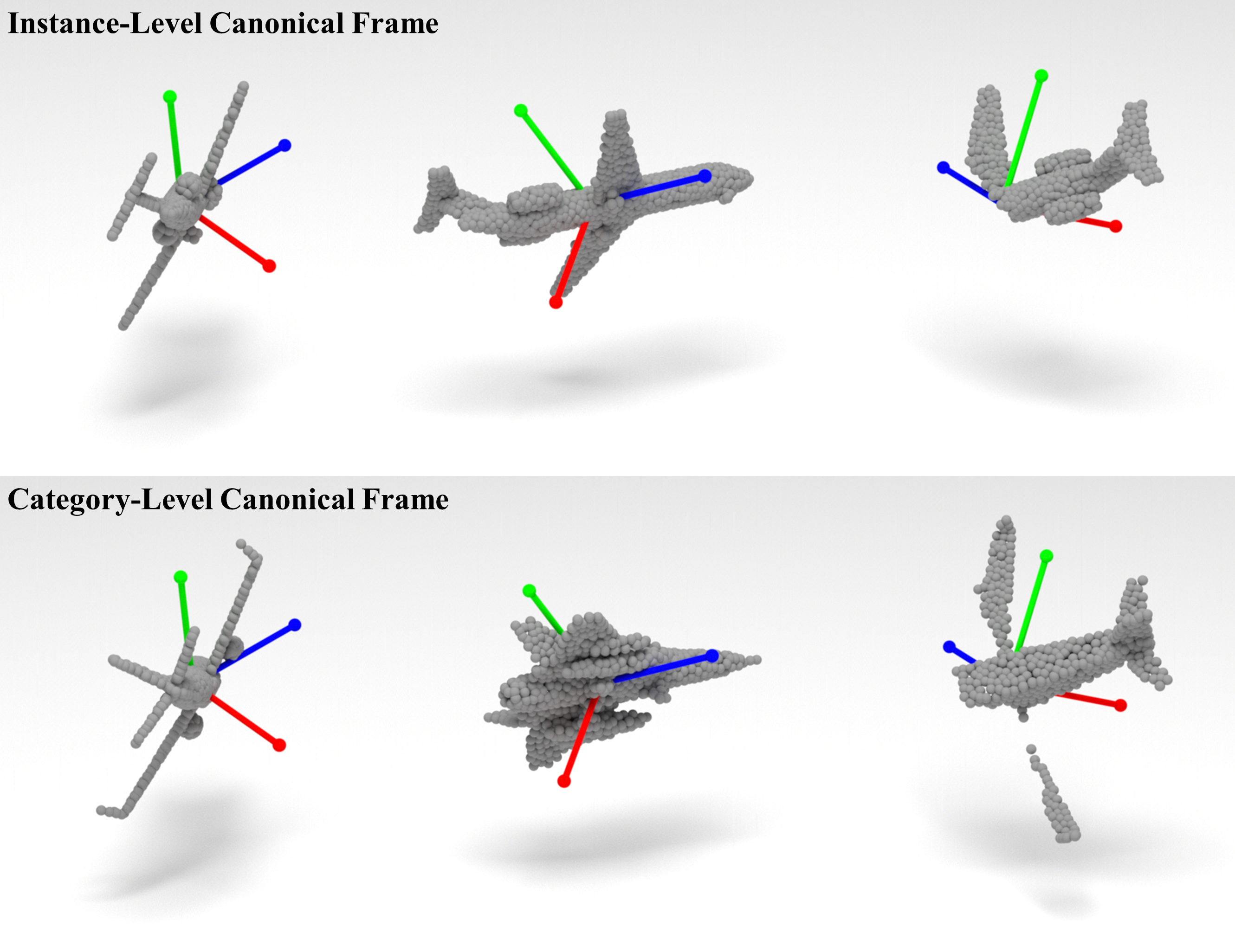}
  \vspace{-0.4in}
  \caption{A canonical frame visualized for (\emph{\textbf{top}})~the same instance in different 3D poses, and (\emph{\textbf{bottom}}) different instances in different 3D poses. Partial shapes with amodal frame shown in last column.
  }
  \vspace{-0.19in}
  \label{fig:pose_canon}
\end{figure}

In \textbf{instance-level 3D pose canonicalization}, our goal is to find a consistent canonical frame across different poses of the same object instance (\cref{fig:pose_canon}, top).
%
%
In \textbf{category-level 3D pose canonicalization}, we want a canonical frame that is consistent with respect to the geometry and local shape across different object instances (\cref{fig:pose_canon}, bottom).
Any equivariant frame that is consistent across shapes
is a valid canonical frame -- this allows us to compare canonicalization with manually-labeled ground truth (see \cref{sec:canonicalization_metrics}).
In addition to full shapes, we also consider partial shape canonicalization for which we define an \emph{amodal} canonical frame as shown in \cref{fig:pose_canon}.



\parahead{Tensor Field Networks\label{sec:tfn}}
Our method estimates a canonical frame for 3D shapes represented as point clouds.
For this task, we use Tensor Field Networks \cite{thomas2018tensor} (TFNs), a 3D architecture that is equivariant to point permutation and rotation, and invariant to translation.
Given a point cloud $X \in \mathbb{R}^{3 \times K}$ and a integer (aka type) $\ell \in \mathbb{N}$, a TFN can produce global (type $\ell$) feature vectors of dimension $2\ell + 1$ stacked in a matrix $F^{\ell} \in \mathbb{R}^{(2\ell + 1) \times C}$, where $C$ is user-defined number of channels.
$F_{:,j}^{\ell}(X)$  satisfies the equivariance property
%
    $F_{:,j}^{\ell}(RX) = D^{\ell}(R)F_{:,j}^{\ell}(X)$,
%
where $D^{\ell}: \mathrm{SO}(3) \rightarrow \mathrm{SO}(2\ell + 1)$ is the so-called Wigner matrix (of type $\ell$) \cite{chirikjian2001engineering, knapp2001representation, lang2020wigner}. 
Please see \cite{thomas2018tensor, anderson2019cormorant, weiler20183d, poulenard2021functional} for details.

\section{Method}
\label{sec:method}
\begin{figure*}[ht!]
\centering
  \includegraphics[width=\textwidth]{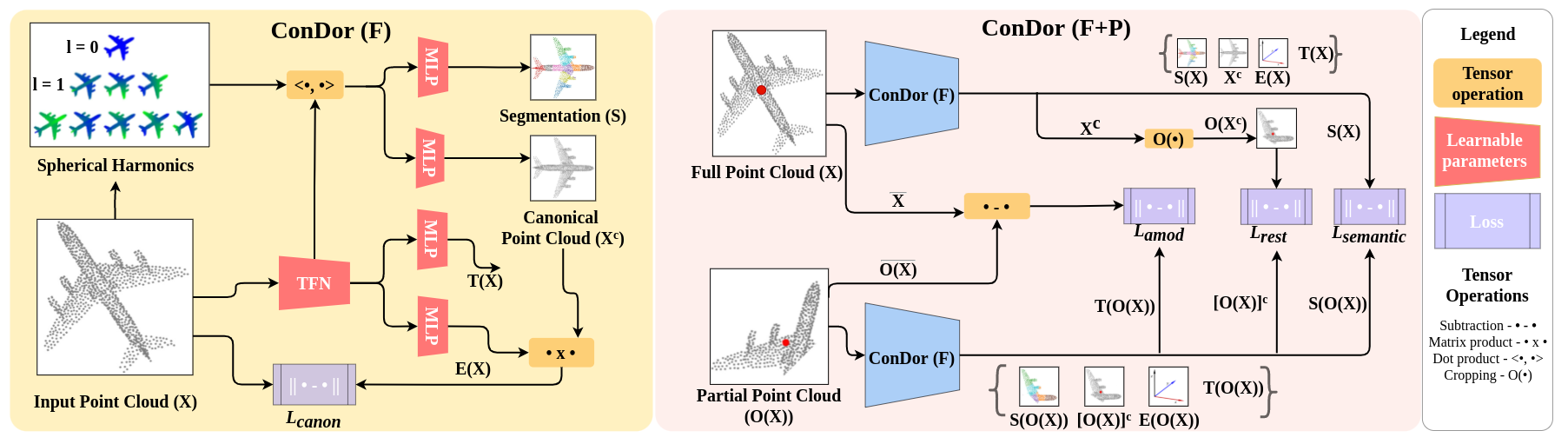}
  \vspace{-0.3in}
  \caption{\textbf{ConDor}. (\emph{\textbf{left}})~Our method learns to canonicalize rotation by estimating an equivariant pose $E(X)$ and an invariant point cloud $X^c$ of an input shape $X$.
  A self-supervision loss ensures that the input and transformed canonical shapes match.
(\emph{\textbf{right}})~To handle translation in partial shapes, we train a two-branch (Siamese) architecture, one taking the full shape and the other taking an occluded (\eg,~via slicing) version of the full shape as input.
Various losses ensure that the feature embeddings of the full and partial shapes match.
We predict the amodal barycenter of the full shape $T(\mathcal{O(X)})$ from the partial shape to canonicalize for position.
}
  \label{fig:pipeline}
\vspace{-0.4cm}
\end{figure*}

%
Given a point cloud $X \in \mathbb{R}^{3 \times K}$ denoting a full or partial shape from a set of non-aligned shapes, our goal is to
estimate its rotation $\mathcal{R}(X)$ (canonical frame) sending $X$ to a canonical pose. For a partial shape $Y \subset X$ we also learn a translation $\mathcal{T}(Y)$ aligning $Y$ with $X$ in the canonical frame.
We achieve this by training a neural network on 3D shapes in a self-supervised manner (see \cref{fig:pipeline}).



\setlength{\belowdisplayskip}{1pt} \setlength{\belowdisplayshortskip}{1pt}
\setlength{\abovedisplayskip}{5pt} \setlength{\abovedisplayshortskip}{5pt}

\subsection{Learning to Canonicalize Rotation}
\label{sec:rotation}
We first discuss the case of canonicalizing 3D rotation for full shapes.
Given a point cloud $X$, our approach estimates a rotation-invariant point cloud $X^c$, and an equivariant rotation $E$ that rotates $X^c$ to $X$.
Note that for full shapes, translation can be canonicalized using mean centering~\cite{novotny2019c3dpo}, but this does not hold for partial shapes.

\parahead{Rotation Invariant Point Cloud/Embedding}
To estimate a rotation-invariant point cloud, we build on top of a permutation-, rotation-equivariant and translation-invariant neural network architecture: Tensor Field Networks (TFNs)\cite{thomas2018tensor} with equivariant non-linearities for TFNs~\cite{poulenard2021functional}.
%
Given $X$, we use a TFN~\cite{poulenard2021functional} to produce global \textbf{equivariant features}
$F^{\ell}$, with columns $F^{\ell}_{:,j}$ as described in \cref{sec:tfn}.
%

The central observation of \cite{poulenard2021functional} is that the features $F$ have the same rotation equivariance property as coefficients of spherical functions in the spherical harmonics basis, and can therefore be treated as such.
We exploit this property by embedding the shape using the spherical harmonics basis and using the global TFN features $F$ as coefficients of this embedding.
Since the input to the spherical harmonics embedding and the coefficients rotate together with the input shape, they can be used to define a rotation and translation \textbf{invariant embedding} of the shape.
Formally, let $Y^{\ell}(x) \in \mathbb{R}^{2\ell+1}$ be the vector of degree $\ell$ spherical harmonics which are homogeneous polynomials defined over $\mathbb{R}^3$.
We define a rotation invariant embedding of the shape as the dot products
\begin{equation}
    H^{\ell}_{ij} := \langle F^{\ell}_{:,j}, Y^{\ell}(X_i) \rangle,
    \label{eq:inv_emb}
\end{equation}
where $i$ is an index to a single point on the point cloud, and $j$ is the channel index as in \cref{sec:tfn}. Both sides of the dot product are rotated by the same Wigner rotation matrix when rotating the input pointcloud $X$ making $H$ invariant to rotations of $X$. The input point cloud is mean-centered to achieve invariance to translation.
Note that we can use any functional basis of the form $:x \mapsto (\varphi^r(\Vert x \Vert)Y^{\ell}(x))_{r\ell}$, where $(\varphi^r)_r$ are real valued functions to define $H$.

We use the rotation invariant embedding corresponding to $\ell = 1$ (degree 1) to produce a 3D \textbf{invariant shape} through a linear layer on top of $H$.
Note that degree 1 spherical harmonics are the $x,y,z$ coordinates of the input point cloud since $Y^1(x) = x$.
As we show in \cref{sec:segmentation}, other choices for $\ell$ enable us to learn consistent co-segmentation without supervision.
The 3D rotation invariant shape is given by:
%
\begin{equation}
X^c_{i} := \sum_j W_{:,j}H^1_{ij} = W (F^{1})^\top X_i.
    \label{eq:eq_3d_inv_embedding}
\end{equation}
We obtain our canonical frame as described in \cref{sec:tfn} as $\mathcal{R}(X) = W (F^{1})^{\top}$ where $W$ is the learnable weights matrix of the linear layer.
%


\parahead{Rotation Equivariant Embedding}
Next, we seek to find an equivariant rotation that transforms $X^c$ to $X$.
In addition to the equivariant features $F$, our TFN also outputs a 3D equivariant frame $E$ which we optimise to be a rotation matrix.
$E$ satisfies the equivariance relation $E(R.X) = RE(X)$ so that the point cloud $E(X)X^c$ is rotation equivariant. Note that we could have chosen $E(X) = \mathcal{R}(X)^{\top}$ but we instead choose to learn $E(X)$ independently as this approach generalizes to the case of non-linear embeddings (\eg,~with values other than $\ell = 1$ in \cref{eq:eq_3d_inv_embedding}) which we use for unsupervised segmentation in \cref{sec:segmentation}.

%
Using $E$, we can transform our 3D invariant embedding $X^c$ back to the input equivariant embedding and compare it to the input point cloud.
To handle situations with high occlusion and symmetric objects
we estimate $P$ equivariant rotations and choose the frame that minimizes the $L^2$ norm between corresponding points in the input and the predicted invariant shape.

\subsection{Learning to Canonicalize Translation}
\label{sec:translation}
Next, we discuss canonicalizing 3D translation for \textbf{partial point clouds}, \eg,~acquired from depth sensors or LiDAR.
As noted, translation canonicalization for full shapes is achieved using mean centering~\cite{novotny2019c3dpo}.
Thus, our approach in \cref{sec:rotation} is sufficient for 3D pose canonicalization of full shapes. However, partial shapes can have different centroids depending on how the shape was occluded. 
To address this issue, we extend our approach to additionally find a \textbf{rotation-equivariant translation} $\mathcal{T} \in \mathbb{R}^3$ that estimates the difference between the barycenter of the full and partial shape from the mean-centered partial point cloud that translates it to align with the full shape in the input frame.

In practice, we operationalize the above idea in a two-branch Siamese architecture as shown in \cref{fig:pipeline}.
We slice the input point cloud to introduce synthetic occlusion.
We penalize the network by ensuring semantic consistency between the full and the partial point cloud. Furthermore, our network predicts an amodal translation vector that captures the barycenter of the full shape from the partial input shape.

\subsection{Unsupervised Co-segmentation}
\label{sec:segmentation}
A surprising finding is that our method can be used for unsupervised part co-segmentation~\cite{chen2019bae} of full and partial shapes with little modification.
This result is enabled by finding the rotation invariant embedding $H$ in \cref{eq:inv_emb} corresponding to all $ \ell \geqslant 0$ to produce a \textbf{non-linear invariant embedding}.
To obtain a consistent rotation invariant part segmentation, we segment the input shape into $N$ parts by learning an MLP on top of the rotation invariant embedding.
The part label of each point in the input point cloud is given by
%
$S_i := \mathrm{softmax}[\mathrm{MLP}(H)_i]$.
%
%
Results visualized in the paper include these segmentations as colored labels.
Please see the supplementary material for more details.

\section{Self-Supervised Learning}
\subsection{Loss Functions}
\label{sec:losses}
A key contribution of our work is to demonstrate that 3D pose canonicalization can be achieved through self-supervised learning as opposed to supervised learning from labeled datasets~\cite{shapenet2015,wu20153d}.
We now list the loss functions that enable this.
Additionally, we describe losses that prevent degenerate results, handle symmetric shapes, and enable unsupervised segmentation.
We begin with full shapes.





\parahead{Canonical Shape Loss}
Our primary self-supervision signal comes from the canonical shape loss that tries to minimize the $L^2$ loss between the rotation invariant point cloud $X^c$ transformed by the rotation equivariant rotation $E$ with the input point cloud $X$.
It is worth noting that $X^c$ and $X$ are in correspondence because our method is permutation equivariant and we extract point-wise embeddings.
For each point $i$ in a point cloud of size $K$, we define the canonical shape loss to be

\begin{align}
    \mathcal{L}_{canon} = \frac{1}{K}\sum_i \Vert EX^c_{i} - X_{i}  \Vert_2. 
        \label{eq:l2_loss}
\end{align}
We empirically observe that our estimation of $E$ can be flipped 180$^\circ$ or $X^c$ can become a degenerate shape when the object class has symmetry or heavy occlusions.
To mitigate this issue, we estimate $P$ equivariant rotations $E_p$ and choose the one that minimizes the above loss.

\parahead{Orthonormality Loss}
The equivariant rotation $E$ estimated by our method must be a valid rotation in $\mathrm{SO}(3)$, but this cannot be guaranteed by the TFN.
We therefore add a loss to constrain $E$ to be orthonormal by minimizing its difference to its closest orthonormal matrix.
We achieve this using the SVD decomposition of $E = U\Sigma V^{\top}$ and enforcing unit eigenvalues with the loss
\begin{align}
    \begin{split}
        \mathcal{L}_{ortho} = \Vert UV^{\top} - E\Vert_2.
        \label{eq:orth_loss}
    \end{split}
\end{align}

\parahead{Separation Loss}
When estimating $P$ equivariant rotations $E_p$, our method could learn a degenerate solution where all $E_p$ are similar.
To avoid this problem, we introduce a separation loss that encourages the network to estimate different equivariant rotations as
%
\begin{align}
    \mathcal{L}_{sep} = -\frac{1}{9P}\sum_{i \neq j} \left|\left| E_i - E_j\right|\right|_2.
\end{align}

\parahead{Restriction Loss}
%
%
We next turn our attention to partial shapes.
Similar to full shapes, we compute the canonical shape, orthonormality and separation losses.
We assume that a partial shape is a result of a cropping operator $\mathcal{O}$ that acts on a full point cloud $X$ to select points corresponding to a partial version $\mathcal{O}(X) \subseteq X$.
In practice, our cropping operator is slicing or image projection (see \cref{sec:training}).
During training, we train two branches of our method, one with the full shape and the other with a partial shape generated using a random sampling of $\mathcal{O}$.
We then enforce that the invariant embedding for partial shapes is a restriction of the invariant embedding of the full shape $X$ using the loss
%

%
\vspace{-0.2in}
\begin{align}
    \mathcal{L}_{rest} = \frac{1}{|\mathcal{S}|}\sum_{ i \in \mathcal{S}} \Vert \widehat{\mathcal{O}[X^c]}_i - \left(\widehat{\mathcal{O}[X]}^{c}\right)_i\Vert_2^2,
    \label{eq:frame_restriction_loss}
\end{align}
where $\mathcal{S}$ is the set of valid indices of points in both $X$ and $\mathcal{O}(X)$, and the $\widehat{\textrm{hat}}$ indicates mean-centered point clouds.
During inference, we do not require the full shape and can operate only with partial shapes.
Empirically, we observe that our method generalizes to different cropping operations between training and inference (see \cref{sec:ablation})



\parahead{Amodal Translation Loss}
Finally, to align the mean-centered partial shape with the full shape, we estimate the barycenter of the full shape after the occlusion operation $\overline{\mathcal{O}[X]}$ from the partial shape only using a rotation-equivariant translation vector $\mathcal{T}(\widehat{\mathcal{O}[X]})$ by minimizing
\begin{align}
    \mathcal{L}_{amod} =  \Vert \mathcal{T}(\widehat{\mathcal{O}[X]}) - \overline{\mathcal{O}(X)} \Vert^2_2.
\end{align}

%


\parahead{Unsupervised Part Segmentation Losses}
%
A surprising finding in our method is that we can segment objects into parts consistently across instances without any supervision (see \cref{fig:teaser}).
This is enabled by interpreting higher degree invariant embedding $H^{\ell}$ as a feature for unsupervised segmentation.
Our losses are based on the localization and equilibrium losses of \cite{sun2020canonical}.
We refer the reader to \cite{sun2020canonical} and the supplementary document for details on these losses.
Note that \cite{sun2020canonical} need to perform segmentation to enable rotation canonicalization, while it is optional for us.

\subsection{Network Architecture \& Training}
\label{sec:training}
Our method is trained on a collection of un-canonicalized shapes $\mathcal{X}$, and partial shapes randomly generated using a suitable operator $\mathcal{O}$.
We report two kinds of partiality: slicing and image projection (\ie,~depth maps).
We borrow our TFN architecture from \cite{poulenard2021functional} and use the ReLU non-linearity in all layers.
We use 1024 and 512 points for full and partial point cloud.
Our method predicts 5 canonical frames for every category.
Our models are trained for 45,000 iterations for each category with the Adam~\cite{kingma2014adam} optimizer with an initial learning rate of $6 \times 10^{-4}$.
We set a step learning rate scheduler that decays our learning rate by a factor of $10^{-1}$ every 15,000 steps.
Our models are trained on Linux with Nvidia Titan V GPUs -- more details in the supplementary document.

\section{Experiments}
\label{sec:expt}
We present quantitative and qualitative results to compare our method with baselines and existing methods, justify design choices, and demonstrate applications.

\begin{table*}[!ht]
\centering
    \caption{Full shape canonicalization compared to a PCA baseline, Canonical Capsules (CaCa)~\cite{sun2020canonical} and Compass~\cite{spezialetti2020learning}, and full (F) and full+partial (F+P) versions of our method.
    We outperform methods on most categories and metrics.
    \vspace{-0.1in}
    \label{table:canonicalization_metrics_full}
    }
\scalebox{0.73}{
\begin{tabular}{r|rrrrrrrrrrrrr|r|r}
    \toprule
     & \textbf{bench} & \textbf{cabinet} & \textbf{car} & \textbf{cellph.} & \textbf{chair} & \textbf{couch} & \textbf{firearm} & \textbf{lamp} & \textbf{monitor} & \textbf{plane} & \textbf{speaker} & \textbf{table} & \textbf{water.} & \textbf{avg.} & \textbf{multi} \\
     
     \midrule
    \multicolumn{16}{l}{\textbf{Instance-Level Consistency (IC)} {\color{blue} $\downarrow$}} \\
    
    \midrule
    \multicolumn{1}{l|}{PCA} & 0.0573 & 0.0350 & 0.0477 & 0.0276 & 0.0974 & 0.0628 & 0.0324 & 0.0755 & 0.0480 & 0.0502 & 0.0491 & 0.0727 & 0.0400 & 0.0535 & 0.0535 \\
    \multicolumn{1}{l|}{CaCa \cite{sun2020canonical}} & 0.0630 & 0.1567 & 0.0426 & 0.0823 & 0.0253 & 0.1479 & 0.0084 & \textbf{0.0372} & 0.0748 & \textbf{0.0093} & 0.1540 & 0.0787 & 0.0270 & 0.0698 & 0.0395 \\ 
    \multicolumn{1}{l|}{Compass \cite{spezialetti2020learning}} & 0.1030 & 0.0816 & 0.0790 & 0.0664 & 0.0791 & 0.0766 & 0.0748 & 0.0495 & 0.0638 & 0.0610 & 0.0721 & 0.0641 & 0.0430 & 0.0703 & 0.0507 \\
    \midrule
    \multicolumn{1}{l|}{Ours (F)} & \textbf{0.0225} & 0.0346 &  \textbf{0.0191} & \textbf{0.0234} & \textbf{0.0221} & \textbf{0.0221} & \textbf{0.0081} & 0.0454 & 0.0283 & 0.0163 & 0.0787 & 0.0523 & 0.0270 & \textbf{0.0308} & 0.0394 \\ 
    \multicolumn{1}{l|}{Ours (F+P)} & 0.0696 & \textbf{0.0288} &  0.0230 & 0.0263 & 0.0235 & 0.0222 & 0.0084 & 0.0403 & \textbf{0.0242} & 0.0144 & \textbf{0.0678} & \textbf{0.0361} & \textbf{0.0236} & 0.0314 & \textbf{0.0329} \\ 
    \midrule
    
    \multicolumn{16}{l}{\textbf{Category-Level Consistency (CC)} {\color{blue} $\downarrow$}} \\
    
    \midrule
    \multicolumn{1}{l|}{Ground truth} & 0.0980 & 0.1460 & 0.0578 & 0.0733 & 0.1191 & 0.0955 & 0.0536 & 0.2147 & 0.1088 & 0.0673 & 0.1709 & 0.1444 & 0.0915 & 0.1108 & 0.1108 \\
    \multicolumn{1}{l|}{PCA} & \textbf{0.0976} & \textbf{0.1055} & 0.0654 & \textbf{0.0600} & 0.1389 & 0.0937 & 0.0527 & 0.1802 & \textbf{0.0970} & 0.0731 & \textbf{0.1397} & 0.1479 & \textbf{0.0816} & 0.1026 & 0.1026 \\
    \multicolumn{1}{l|}{CaCa \cite{sun2020canonical}} & 0.1134 & 0.1742 & 0.0730 & 0.1033 & 0.1220 & 0.1919 & \textbf{0.0493} & 0.1888 & 0.1186 & 0.0684 & 0.1840 & 0.1660 & 0.0883 & 0.1262 & 0.1132 \\ 
    \multicolumn{1}{l|}{Compass \cite{spezialetti2020learning}} & 0.1654 & 0.1348 & 0.1077 & 0.0931 & 0.1522 & 0.1175 & 0.1258 & 0.1833 & 0.1266 & 0.1019 & 0.1579 & 0.1626 & 0.0942 & 0.1325 & 0.1283 \\
    \midrule
    \multicolumn{1}{l|}{Ours (F)} & 0.1043 & 0.1067 & \textbf{0.0575} & 0.0612 & \textbf{0.1135} & \textbf{0.0869} & 0.0525 & \textbf{0.1754} & 0.0988 & \textbf{0.0681} & 0.1504 & 0.1475 & 0.0851 & \textbf{0.1006} & 0.1035 \\
    \multicolumn{1}{l|}{Ours (F+P)} & 0.1250 & 0.1065 &  0.0581 & 0.0635 & 0.1145 & 0.0874 & 0.0500 & 0.1844 & 0.1001 & 0.0679 & 0.1477 & \textbf{0.1432} & 0.0912 & 0.1030 & \textbf{0.1005} \\
    \midrule
    
    \multicolumn{16}{l}{\textbf{Ground Truth Consistency (GC)}{\color{blue} $\downarrow$}} \\ 
    
    \midrule
    \multicolumn{1}{l|}{PCA} & 0.0760 & 0.1047 & \textbf{0.0208} & \textbf{0.0390} & 0.1190 & 0.0799 & 0.0261 & 0.1366 & 0.0862 & 0.0460 & 0.1280 & 0.1267 & 0.0645 & 0.0810 & \textbf{0.0810} \\
    \multicolumn{1}{l|}{CaCa~\cite{sun2020canonical}} & 0.0761 & \textbf{0.0688} & 0.0529 & 0.0667 & 0.0943 & 0.1812 & 0.0330 & 0.1592 & 0.0897 & 0.0266 & \textbf{0.0744} & 0.1401 & 0.0683 & 0.0870 & 0.1060 \\ 
    \multicolumn{1}{l|}{Compass \cite{spezialetti2020learning}} & 0.1599 & 0.1586 & 0.0892 & 0.0851 & 0.1504 & 0.1160 & 0.1214 & 0.1654 & 0.1231 & 0.0975 & 0.1552 & 0.1554 & 0.0804 & 0.1275 & 0.1247 \\
    \midrule
    \multicolumn{1}{l|}{Ours (F)} & \textbf{0.0671} & 0.1131 & 0.0257 & 0.0511 & 0.0526 & 0.0585 & 0.0359 & 0.1399 & 0.0674 & \textbf{0.0255} & 0.1505 & 0.0779 & 0.0746 & 0.0723 & 0.0902 \\
    \multicolumn{1}{l|}{Ours (F+P)} & 0.1115 & 0.1134 & 0.0230 & 0.0553 & \textbf{0.0509} & \textbf{0.0537} & \textbf{0.0223} & \textbf{0.1274} & \textbf{0.0650} & 0.0286 & 0.1456 & \textbf{0.0738} & \textbf{0.0477} & \textbf{0.0706} & 0.0843 \\
    \bottomrule
\end{tabular}}
\vspace{-0.5cm}
\end{table*}

\parahead{Datasets}
For full shapes, we use un-canonicalized shapes from ShapeNet (Core)~\cite{chang2015shapenet} and ModelNet40~\cite{wu20153d}.
For ShapeNet, our data split~\cite{deprelle2019learning, sun2020canonical} has 31,747 train shapes, and 7,943 validation shapes where each shape is a 3D point cloud with 1024 points sampled using farthest point sampling.
The shapes are from 13 classes: airplane, bench, cabinet, car, chair, monitor, lamp, speaker, firearm, couch, table, cellphone, and watercraft.
For ModelNet40~\cite{wu20153d}, we use 40 categories with 12,311 shapes (2,468 test).
For partial shapes, we either randomly slice shapes from the above datasets, or we use the more challenging ShapeNetCOCO dataset~\cite{sridhar2019multiview} that contains 
objects viewed from multiple camera angles and mimics occlusions from depth sensors.
While all these datasets are already pre-canonicalized, we use this information \textbf{only for evaluation} -- our method is trained on randomly transformed un-canonicalized shapes $X \in \mathcal{X}$ from these datasets.


\subsection{Canonicalization Metrics}
\label{sec:canonicalization_metrics}
Most work on canonicalization evaluates performance indirectly on downstream tasks such as segmentation or registration~\cite{sun2020canonical, spezialetti2020learning}.
This makes it hard to disentangle canonicalization performance from task performance.
We contribute four new metrics that measure different aspects of 3D pose canonicalization while disentangling performance from downstream tasks.
The first three of these metrics evaluate rotation assuming mean-centering, while the last metric measures translation errors for partial shapes.

\parahead{Instance-Level Consistency (IC)}
%
The IC metric is designed to evaluate how well a method performs for canonicalizing the 3D rotation of the \emph{same shape instance}.
For each shape in the dataset, we obtain another copy of it by applying a rotation from $\mathbf{R}$, a user-defined set of random rotations (we use 120 rotations).
We then compute the 2-way Chamfer Distance ($\mathrm{CD}$), to handle classes with symmetries such as tables, between the canonicalized versions of the shapes (with superscript $^c$).
We expect this to be as small as possible for better canonicalization.
The average IC metric is given as:
%
\begin{align}
    \mathrm{IC} := \frac{1}{|\mathcal{X}||\mathbf{R}|} \sum_{X_i \in \mathcal{X}} \sum_{R_j \in \mathbf{R}}\mathrm{CD}[ (R_j.X_i)^c, X^c ].\nonumber
\end{align}

\parahead{Category-Level Consistency (CC)}
The CC metric is designed to evaluate the quality of 3D rotation canonicalization between \emph{different shape instances}.
For each shape $X$ in the dataset, we pick $N$ other shapes to form a set of comparison shapes $\mathcal{N}$.
We then follow a similar approach as $\mathrm{IC}$ and compute the 2-way Chamfer Distance between each shape and its $N$ possible comparison shapes.
Intuitively, we expect this metric to be low if canonicalization is consistent across different instances.
Ideally, we want to evaluate this metric for all possible comparison shapes, but to reduce computation time, we pick $N = 120$ random comparison shapes.
The average CC metric is given as:
%
\begin{align}
    \mathrm{CC} := \frac{1}{|\mathcal{X}| N} \sum_{X_i \in \mathcal{X}} \sum_{X_j \in \mathcal{N}} \mathrm{CD}[ X_i^c , X_j^c ].\nonumber
\end{align}

\parahead{Ground Truth Consistency (GC)}
The GC metric is designed to compare estimated canonicalization with manual ground truth pre-canonicalization in datasets like ShapeNet and ModelNet40. For perfect canonicalization, the predicted canonical shape should be a constant rotation away from ground truth shape. Given the predicted canonicalizing frames $\mathcal{R}(X_j), \mathcal{R}(X_k)$ for aligned shapes $X_j, X_k \in \mathcal{X}$, we induce the same canonicalization on any other shape $X_i \in \mathcal{X}$ and compute the 2-way $\mathrm{CD}$ between them.
%
\begin{align}
     \mathrm{GC} :=
     \frac{1}{|\mathcal{X}|^3}\sum_{X_i,X_j,X_k \in \mathcal{X}} \hspace{-3mm} \mathrm{CD}[\mathcal{R}(X_j).X_i, \mathcal{R}(X_k).X_i].\nonumber
 \end{align}




%
We note that manual canonicalization, which is based on human semantic understanding of shapes, does not necessarily match with this paper's notion of canonicalization which is founded on geometric similarity.
Nonetheless, this metric provides a way to compare with human annotations.


\parahead{Translation Error (TE)}
%
%
To measure error in translation for partial shapes, we compute the average $L^2$ norm between the estimated amodal translation and ground truth amodal translation -- this has the same form as $\mathcal{L}_{amod}$ in \cref{sec:losses}.
Note that we have the ground truth amodal translation for our datasets since partial shapes are generated from the full shapes using an occlusion function $\mathcal{O}$.

\begin{table*}[!ht]
\centering
\caption{Partial shape canonicalization compared to PCA and Compass*, our modification of \cite{spezialetti2020learning}. We outperform other methods by a larger margin than in the full shapes setting.
    \label{table:canonicalization_metrics_partial}}
    \vspace{-0.1in}
\scalebox{0.73}{
\begin{tabular}{r|rrrrrrrrrrrrr|r|r}
    \toprule
     & \textbf{bench} & \textbf{cabinet} & \textbf{car} & \textbf{cellph.} & \textbf{chair} & \textbf{couch} & \textbf{firearm} & \textbf{lamp} & \textbf{monitor} & \textbf{plane} & \textbf{speaker} & \textbf{table} & \textbf{water.} & \textbf{avg.} & \textbf{multi} \\
    \midrule

    \multicolumn{16}{l}{\textbf{Ground Truth Consistency (GC)}{\color{blue} $\downarrow$}} \\
    \midrule
    
    \multicolumn{1}{l|}{PCA} & \textbf{0.0916} & 0.1391 & 0.0727 & 0.0879 & 0.1337 & 0.0908 & 0.0371 & 0.1985 & 0.0804 & 0.0915 & 0.1479 & 0.1087 & 0.1021 & 0.1063 & 0.1063 \\
    \multicolumn{1}{l|}{Compass*} & 0.1917 & 0.1412 & 0.1020 & 0.1066 & 0.1476 & 0.1115 & 0.1538 & 0.1735 & 0.1194 & 0.1115 & 0.1617 & 0.1709 & \textbf{0.0737} & 0.1358 & 0.1423 \\
    \multicolumn{1}{l|}{Ours(F+P)} & 0.1416 & \textbf{0.1182} & \textbf{0.0356} & \textbf{0.0685} & \textbf{0.0780} & \textbf{0.0593} & \textbf{0.0300} & \textbf{0.1501} & \textbf{0.0692} & \textbf{0.0360} & \textbf{0.1469} & \textbf{0.0662} & 0.0739 & \textbf{0.0826} & \textbf{0.1016} \\
    \midrule
    
    \multicolumn{16}{l}{\textbf{Instance-Level Consistency (IC)} {\color{blue} $\downarrow$}} \\
    \midrule
    \multicolumn{1}{l|}{PCA} &  \textbf{0.1033} & 0.1140 & 0.1149 & 0.0828 & 0.1475 & 0.1221 & 0.0517 & 0.1571 & 0.0867 & 0.1000 & 0.1182 & 0.1401 & 0.0756 & 0.1088 & 0.1088 \\
    \multicolumn{1}{l|}{Compass*} & 0.1900 & 0.0790 & 0.1183 & 0.0911 & 0.1280 & 0.1053 & 0.1440 & 0.1000 & 0.0836 & 0.1000 & 0.1134 & 0.1080 & 0.0487 & 0.1084 & 0.1247 \\
    \multicolumn{1}{l|}{Ours(F+P)} & 0.1432 & \textbf{0.0501} & \textbf{0.0349} & \textbf{0.0442} & \textbf{0.0622} & \textbf{0.0478} & \textbf{0.0221} & \textbf{0.0891} & \textbf{0.0442} & \textbf{0.0265} & \textbf{0.1086} & \textbf{0.0739} & \textbf{0.0469} & \textbf{0.0611} & \textbf{0.0792} \\
    \midrule
    
    \multicolumn{16}{l}{\textbf{Category-Level Consistency (CC)} {\color{blue} $\downarrow$}} \\
    \midrule
    
    \multicolumn{1}{l|}{PCA} &  \textbf{0.1269} & 0.1500 & 0.1253 & 0.1081 & 0.1636 & 0.1367 & 0.0691 & 0.2312 & 0.1178 & 0.1124 & 0.1677 & 0.1769 & 0.1078 &  0.1380 & 0.1380 \\ 
    \multicolumn{1}{l|}{Compass*} &  0.2118 & 0.1300 & 0.1438 & 0.1215 & 0.1612 & 0.1280 & 0.1688 & \textbf{0.1990} & 0.1242 & 0.1255 & 0.1760 & 0.1719 & \textbf{0.0919} &  0.1503 & 0.1647 \\
    \multicolumn{1}{l|}{Ours (F+P)} &  0.1695 & \textbf{0.1109} & \textbf{0.0632} & \textbf{0.0739} & \textbf{0.1270} & \textbf{0.0935} & \textbf{0.0546} & 0.2048 & \textbf{0.1042} & \textbf{0.0713} & \textbf{0.1666} & \textbf{0.1579} & 0.0936 & \textbf{0.1147} & \textbf{0.1234} \\
    
   \bottomrule
\end{tabular}}
\vspace{-0.5cm}
\end{table*}

\subsection{Comparisons}
\label{sec:evaluation_full}
We report comparisons on canonicalizing both full and partial shapes.
Only the rotation metrics from \cref{sec:canonicalization_metrics} are relevant for full shapes since we assume input shapes are mean-centered without translation differences~\cite{novotny2019c3dpo}.
We report the TE metric for partial shape canonicalization.
Outside of these metrics, we also report indirect evaluations of canonicalization~\cite{sun2020canonical, spezialetti2020learning} on classification.

\parahead{Canonicalization Metrics}
We compare our method with baselines and other methods using our new canonicalization metrics (\cref{sec:canonicalization_metrics}).
For this experiment, we follow previous work~\cite{deprelle2019learning} and choose 13 categories from the ShapeNet, training one model per category as well as a joint model for all categories.
We choose PCA as a baseline -- for each shape we compute the top-3 principal components and use this as an equivariant frame for alignment across instances.
We compare with two methods for rotation canonicalization: Canonical Capsules (CaCa)~\cite{sun2020canonical} and Compass~\cite{spezialetti2020learning}.


Results for full shape canonicalization are shown in \cref{table:canonicalization_metrics_full}. 
We evaluate two versions of our method on full shapes, one trained with only full shapes (F) and one trained on both full and partial shapes (F+P).
For the IC metric, both our methods outperform other methods, including baselines, in almost all categories. PCA underperforms in the IC metric due to the frame ambiguity.  
Our method outperforms other canonicalization methods, but surprisingly, we find that PCA is very close. For the CC metric, canonicalized shapes of different geometry are compared with each other. PCA minimizes CC metric by aligning shapes using the principal directions, but does not result in the correct canonical frame as shown in \cref{fig:money_shot} (see supplement for in-depth discussion).
Qualitative results in \cref{fig:money_shot} show that we perform significantly better than other methods.
Finally, our method outperforms other methods on the GC metric indicating that it could be used to extend the size of existing datasets (see \cref{sec:applications}).

Next, we discuss results of partial shape canonicalization shown in \cref{table:canonicalization_metrics_partial}. 
Since no other method exists for partial shape canonicalization, we modified the training setting of Compass to include slicing augmentation (using $\mathcal{O}$) to operate similar to our F+P method (Compass*).
The training data and occlusion function are identical for all methods. 
Different from full shapes, we observe that our method significantly outperforms other methods on all three metrics indicating that our method's design is suited for handling partiality.
We also compute the Translation Error (TE) metric averaged over all our single category models as \textbf{0.0291} while it is \textbf{0.0326} for our multi-category model.
For comparison, all our shapes lie within a unit-diagonal cuboid~\cite{shapenet2015}.
%
\begin{figure*}[t!]
\centering
  \label{fig:money_shot}
  \includegraphics[width=\textwidth]{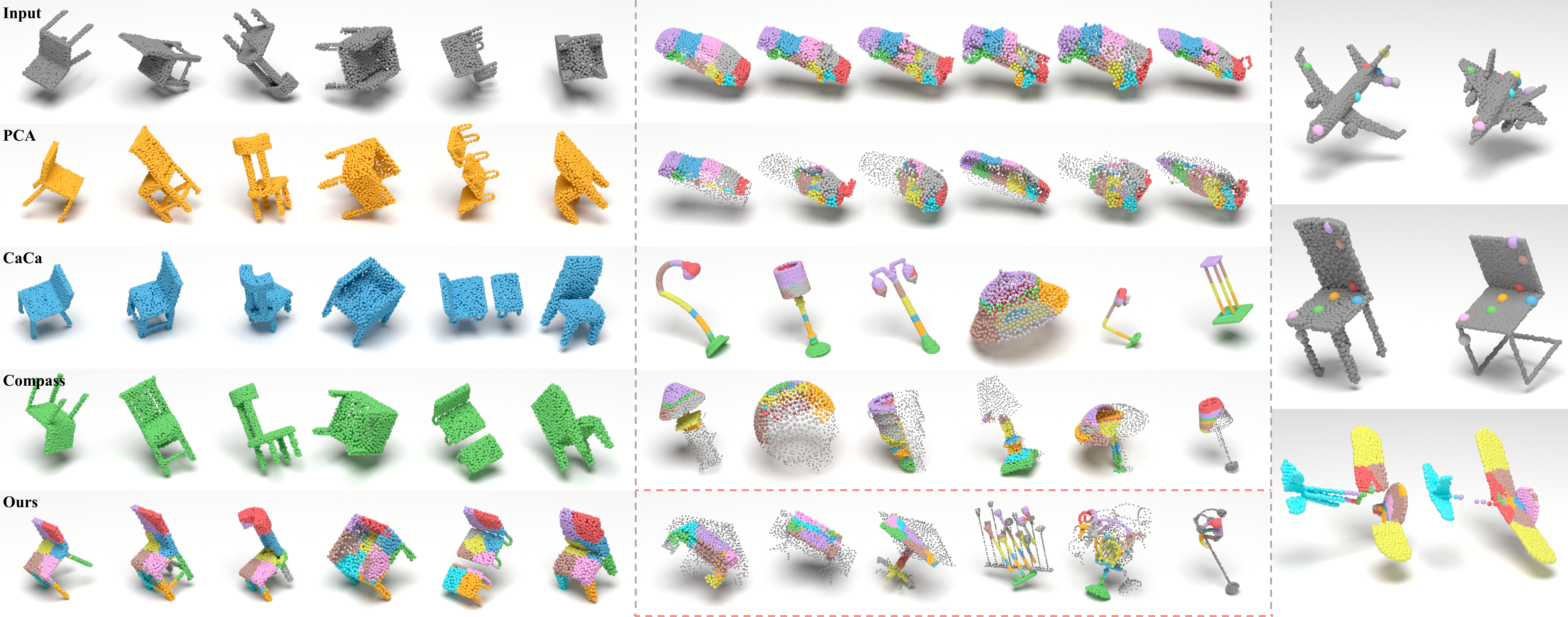}
  \vspace{-0.3in}
  \caption{(\emph{\textbf{left}})~Qualitative comparison with other methods on 6 \emph{randomly chosen} full shapes.
  (\emph{\textbf{center}})~More qualitative results from our method on challenging full/partial car shapes and a variety of full/partial lamp shapes (missing parts only shown for visualization).
  The last row (red border) shows failure cases caused due to incorrect canonical translation for partial shapes, or symmetric shapes.
  (\emph{\textbf{right}})~Rows 1--2: Application of our method in transferring sparse keypoints from one shape to another.
  Row 3: Canonicalization of two depth maps from the ShapeNetCoco~\cite{sridhar2019multiview} dataset showing consistency in canonicalized shapes.
  All results rendered using Mitsuba 2~\cite{NimierDavidVicini2019Mitsuba2}.
  }
  \vspace{-0.1in}
\end{figure*}



\parahead{3D Shape Classification}
We measure 3D shape classification accuracy as an indirect metric of canonicalization following~\cite{spezialetti2020learning}.
We train models with un-canonicalized shapes from all 13 categories.
We augment the PCA baseline, CaCa, Compass and our full shape models with PointNet~\cite{qi2017pointnet} which performs classification on canonicalized outputs.
We observe that our method (\textbf{74.6\%}) outperforms other methods on classification accuracy: PCA
(64.9\%), CaCa (72.5\%), and Compass (72.2\%).
Please see the supplementary document for comparison on registration.

\parahead{Registration}
We measure the registration accuracy of our method for categories (airplanes, chairs, multi) on full shapes in table~\ref{table:registration_shapenet}. Our method does not perform well in this task as we predict a frame $E \in O(3)$ which can have reflection symmetries resulting in high $\mathrm{RMSE}$, but low $\mathrm{CD}$.

\begin{table}[H]
\begin{center}
\vspace{-0.3cm}
\caption{\textbf{Registration} – Distance in terms of root mean-square error (RMSE) and Chamfer distance between registered and ground-truth points on the ShapeNet (core) dataset for full shapes only. 
\label{table:registration_shapenet}}
\scalebox{0.65}{
\begin{tabular}{l|ccc|ccc}
\toprule
\textbf{}                    & \multicolumn{3}{l|}{\textbf{RMSE}{\color{blue} $\downarrow$}}                  & \multicolumn{3}{l}{\textbf{Chamfer (CD)}{\color{blue} $\downarrow$}}                \\
\midrule
\textbf{Method}              & \textbf{Airplane} & \textbf{Chair} & \textbf{Multi} & \textbf{Airplane} & \textbf{Chair} & \textbf{Multi} \\
\midrule
PCA                 & 0.616             & 0.695          & 0.715          & 0.050             & 0.097          & 0.054          \\
Deep Closest Points \cite{wang2019deep}  & 0.318             & 0.160          & 0.131          & -                 & -              & -              \\
Deep GMR \cite{yuan2020deepgmr}            & 0.079             & 0.082          & 0.077          & -                 & -              & -              \\
CaCa \cite{sun2020canonical}                & \textbf{0.024}             & \textbf{0.027}          & \textbf{0.070}          & \textbf{0.009}             & 0.026          & 0.040          \\
Compass \cite{spezialetti2020learning}             & 0.361             & 0.369          & 0.487          & 0.061             & 0.079          & 0.051          \\
Ours (F)            & 0.254             & 0.314          & 0.496          & 0.015             & 0.026          & 0.040          \\
Ours (F + P)        & 0.201             & 0.280          & 0.404          & 0.014             & \textbf{0.023}          & \textbf{0.033} \\  
\bottomrule
\end{tabular}}
\end{center}
\vspace{-0.6cm}
\end{table}

\subsection{Ablations}
\label{sec:ablation}
%
We justify the following key design choices: the effect
of increasing amounts of occlusion/partiality, loss
functions (\cref{sec:losses}), and the benefit of multiple frames.
%

\parahead{Degree of Occlusion/Partiality}
We examine the ability of our model to handle varying amounts of occlusion/partiality for the car category.
Our occlusion function $\mathcal{O}$ occludes shapes to only keep a fraction of the original shape between 25\% and 75\% (\ie,~25\% is more occluded than 75\%).
The average over all metrics indicates that our method performs optimally when trained at 50\% occlusion (25\%: 0.0594, 50\%: \textbf{0.0580}, 75\%: 0.0886).
%

\parahead{Loss Functions}
We evaluate our F+P model on both full
and partial shapes trained
with all losses, without the separation loss $\mathcal{L}_{sep}$, and without the restriction loss $\mathcal{L}_{rest}$. We observe that using $\mathcal{L}_{sep}$ and $\mathcal{L}_{rest}$ performs optimally with the least average error $\textbf{0.0696}$ across all canonicalization metrics over three categories (airplanes, tables, chairs).

\parahead{Multi-Frame Prediction}
We ablate on the number of canonical frames ($1, 3, 5$) predicted by our method to measure its effectiveness on symmetric categories.
We evaluate on two symmetric categories, table, and lamp, and observe (\cref{table:multi_frame_ablation}) that 3 and 5 frames perform better in most cases.
\begin{table}[!h]
\centering
\vspace{-0.1cm}
\caption{Our method handles symmetric categories like lamp and table by estimating multiple canonical frames.
\label{table:multi_frame_ablation}}
\vspace{-0.1in}
\scalebox{0.75}{
\begin{tabular}{c|ccc|ccc}
\toprule
\textbf{Category}       
& \multicolumn{3}{c|}{\textbf{lamp}}        & \multicolumn{3}{c}{\textbf{table}}            \\ \midrule
\textbf{Frames}         
& \textbf{1} & \textbf{3} & \textbf{5}      & \textbf{1} & \textbf{3}      & \textbf{5}      \\ \midrule
\multicolumn{1}{l|}{GC (full) {\color{blue} $\downarrow$}}     & 0.1400     & 0.1370     & \textbf{0.1274} & 0.0749     & \textbf{0.0693} & 0.0738          \\ 
\multicolumn{1}{l|}{IC (full) {\color{blue} $\downarrow$}}      & 0.0686     & 0.0635     & \textbf{0.0403} & 0.0607     & 0.0564          & \textbf{0.0361} \\ 
\multicolumn{1}{l|}{CC (full) {\color{blue} $\downarrow$}}    & 0.1869     & 0.1887     & \textbf{0.1844} & 0.1595     & 0.1569          & \textbf{0.1432} \\ 
\midrule
\multicolumn{1}{l|}{GC (partial) {\color{blue} $\downarrow$}}  & 0.1782     & 0.1711     & \textbf{0.1501} & 0.0681     & \textbf{0.0635} & 0.0662          \\ 
\multicolumn{1}{l|}{IC (partial) {\color{blue} $\downarrow$}}   & 0.1376     & 0.1319     & \textbf{0.0891} & 0.0923     & 0.0936          & \textbf{0.0739} \\ 
\multicolumn{1}{l|}{CC (partial) {\color{blue} $\downarrow$}} & 0.2230     & 0.2226     & \textbf{0.2048} & 0.1705     & 0.1722          & \textbf{0.1579} \\ \bottomrule
\end{tabular}
}
\vspace{-0.5cm}
\end{table}


\subsection{Applications}
\label{sec:applications}
\ShortName enables applications that were previously difficult, particularly for category-level object understanding.
First, since our method operates on partial shapes, we can canonicalize objects in \textbf{depth images}.
%
To validate this, we use depth maps from the ShapeNetCOCO dataset~\cite{sridhar2019multiview} and canonicalize partial point clouds from the depth maps.
\cref{fig:money_shot} (right, row 3) shows an example of depth map canonicalization (see supplementary).
Second, since our method outperforms other methods, we believe it can be used to expand existing canonical datasets with un-canonicalized shapes from the internet -- we show examples of expanding the ShapeNet in the supplementary document.
Finally, we show that \ShortName can be used to transfer sparse keypoint annotations between shape instances.
%
%
We utilize the unsupervised part segmentation learned using our method to solve this task (see supplementary).
\cref{fig:money_shot} (right, rows 1--2) shows results of transferring keypoint annotations from one shape to another.

    

\section{Conclusion}
%

%
We introduced \ShortName, a self-supervised method to canonicalize the 3D pose of full and partial 3D shapes.
Our method uses TFNs and self-supervision losses to learn to canonicalize pose from an un-canonicalized shape collection.
Additionally, we can learn to consistently co-segment object parts without supervision. We reported detailed experiments using four new metrics, and new applications.

\parahead{Limitations \& Future Work}
Despite the high quality of our results, we encounter failures (see \cref{fig:money_shot}), primarily with symmetric or objects with fine details (lamps) where the canonical frame is incorrect. We also observed that PCA often performs very well, and sometimes outperforms methods on full shapes (we do significantly better on partial shapes). Our method occasionally generates flipped canonicalized shapes along the axis of symmetry due to the prediction of an $O(3)$ frame.
Our work can be extended to canonicalize purely from partial shapes and perform scale canonicalization.

\parahead{Acknowledgments}
This work was supported by AFOSR grant FA9550-21-1-0214, a Google Research Scholar Award, a Vannevar Bush Faculty Fellowship, ARL grant W911NF2120104, and gifts from the Adobe and Autodesk corporations. We thank the reviewers for their valuable comments.
\vfil





{\small
\bibliographystyle{ieee_fullname}
\bibliography{ms}

\begin{thebibliography}{10}\itemsep=-1pt

\bibitem{UnityAss22:online}
Unity asset store - the best assets for game making.
\newblock \url{https://assetstore.unity.com/}.
\newblock (Accessed on 11/08/2021).

\bibitem{anderson2019cormorant}
Brandon Anderson, Truong-Son Hy, and Risi Kondor.
\newblock Cormorant: Covariant molecular neural networks.
\newblock {\em arXiv preprint arXiv:1906.04015}, 2019.

\bibitem{chang2015shapenet}
Angel~X Chang, Thomas Funkhouser, Leonidas Guibas, Pat Hanrahan, Qixing Huang,
  Zimo Li, Silvio Savarese, Manolis Savva, Shuran Song, Hao Su, et~al.
\newblock Shapenet: An information-rich 3d model repository.
\newblock {\em arXiv preprint arXiv:1512.03012}, 2015.

\bibitem{shapenet2015}
Angel~X. Chang, Thomas Funkhouser, Leonidas Guibas, Pat Hanrahan, Qixing Huang,
  Zimo Li, Silvio Savarese, Manolis Savva, Shuran Song, Hao Su, Jianxiong Xiao,
  Li Yi, and Fisher Yu.
\newblock {ShapeNet: An Information-Rich 3D Model Repository}.
\newblock Technical Report arXiv:1512.03012 [cs.GR], Stanford University ---
  Princeton University --- Toyota Technological Institute at Chicago, 2015.

\bibitem{chen2019clusternet}
Chao Chen, Guanbin Li, Ruijia Xu, Tianshui Chen, Meng Wang, and Liang Lin.
\newblock Clusternet: Deep hierarchical cluster network with rigorously
  rotation-invariant representation for point cloud analysis.
\newblock In {\em Proceedings of the IEEE Conference on Computer Vision and
  Pattern Recognition}, pages 4994--5002, 2019.

\bibitem{chen2019bae}
Zhiqin Chen, Kangxue Yin, Matthew Fisher, Siddhartha Chaudhuri, and Hao Zhang.
\newblock Bae-net: branched autoencoder for shape co-segmentation.
\newblock In {\em Proceedings of the IEEE/CVF International Conference on
  Computer Vision}, pages 8490--8499, 2019.

\bibitem{chirikjian2001engineering}
Gregory~S Chirikjian, Alexander~B Kyatkin, and AC Buckingham.
\newblock Engineering applications of noncommutative harmonic analysis: with
  emphasis on rotation and motion groups.
\newblock {\em Appl. Mech. Rev.}, 54(6):B97--B98, 2001.

\bibitem{choy20163d}
Christopher~B Choy, Danfei Xu, JunYoung Gwak, Kevin Chen, and Silvio Savarese.
\newblock 3d-r2n2: A unified approach for single and multi-view 3d object
  reconstruction.
\newblock In {\em European conference on computer vision}, pages 628--644.
  Springer, 2016.

\bibitem{cohen2018spherical}
Taco~S Cohen, Mario Geiger, Jonas K{\"o}hler, and Max Welling.
\newblock Spherical cnns.
\newblock {\em arXiv preprint arXiv:1801.10130}, 2018.

\bibitem{cutzu1994canonical}
Florin Cutzu and Shimon Edelman.
\newblock Canonical views in object representation and recognition.
\newblock {\em Vision Research}, 34(22):3037--3056, 1994.

\bibitem{deprelle2019learning}
Theo Deprelle, Thibault Groueix, Matthew Fisher, Vladimir~G Kim, Bryan~C
  Russell, and Mathieu Aubry.
\newblock Learning elementary structures for 3d shape generation and matching.
\newblock {\em arXiv preprint arXiv:1908.04725}, 2019.

\bibitem{esteves2018learning}
Carlos Esteves, Christine Allen-Blanchette, Ameesh Makadia, and Kostas
  Daniilidis.
\newblock Learning so (3) equivariant representations with spherical cnns.
\newblock In {\em Proceedings of the European Conference on Computer Vision
  (ECCV)}, pages 52--68, 2018.

\bibitem{rotpredictor}
Jin Fang, Dingfu Zhou, Xibin Song, Shengze Jin, Ruigang Yang, and Liangjun
  Zhang.
\newblock Rotpredictor: Unsupervised canonical viewpoint learning for point
  cloud classification.
\newblock In {\em 2020 International Conference on 3D Vision (3DV)}, pages
  987--996, 2020.

\bibitem{groueix2018papier}
Thibault Groueix, Matthew Fisher, Vladimir~G Kim, Bryan~C Russell, and Mathieu
  Aubry.
\newblock A papier-m{\^a}ch{\'e} approach to learning 3d surface generation.
\newblock In {\em Proceedings of the IEEE conference on computer vision and
  pattern recognition}, pages 216--224, 2018.

\bibitem{gu2020weaklysupervised}
Jiayuan Gu, Wei-Chiu Ma, Sivabalan Manivasagam, Wenyuan Zeng, Zihao Wang, Yuwen
  Xiong, Hao Su, and Raquel Urtasun.
\newblock Weakly-supervised 3d shape completion in the wild, 2020.

\bibitem{hanocka2019meshcnn}
Rana Hanocka, Amir Hertz, Noa Fish, Raja Giryes, Shachar Fleishman, and Daniel
  Cohen-Or.
\newblock Meshcnn: a network with an edge.
\newblock {\em ACM Transactions on Graphics (TOG)}, 38(4):1--12, 2019.

\bibitem{johnson1997spin}
Andrew~E Johnson.
\newblock Spin-images: a representation for 3-d surface matching.
\newblock 1997.

\bibitem{keskin2013real}
Cem Keskin, Furkan K{\i}ra{\c{c}}, Yunus~Emre Kara, and Lale Akarun.
\newblock Real time hand pose estimation using depth sensors.
\newblock In {\em Consumer depth cameras for computer vision}, pages 119--137.
  Springer, 2013.

\bibitem{kingma2014adam}
Diederik~P Kingma and Jimmy Ba.
\newblock Adam: A method for stochastic optimization.
\newblock {\em arXiv preprint arXiv:1412.6980}, 2014.

\bibitem{knapp2001representation}
Anthony~W Knapp.
\newblock {\em Representation theory of semisimple groups: an overview based on
  examples}, volume~36.
\newblock Princeton university press, 2001.

\bibitem{lang2020wigner}
Leon Lang and Maurice Weiler.
\newblock A wigner-eckart theorem for group equivariant convolution kernels.
\newblock {\em arXiv preprint arXiv:2010.10952}, 2020.

\bibitem{lei2020pix2surf}
Jiahui Lei, Srinath Sridhar, Paul Guerrero, Minhyuk Sung, Niloy Mitra, and
  Leonidas~J Guibas.
\newblock Pix2surf: Learning parametric 3d surface models of objects from
  images.
\newblock In {\em European Conference on Computer Vision}, pages 121--138.
  Springer, 2020.

\bibitem{liu2020fg}
Kangcheng Liu, Zhi Gao, Feng Lin, and Ben~M Chen.
\newblock Fg-net: Fast large-scale lidar point cloudsunderstanding network
  leveraging correlatedfeature mining and geometric-aware modelling.
\newblock {\em arXiv preprint arXiv:2012.09439}, 2020.

\bibitem{maturana2015voxnet}
Daniel Maturana and Sebastian Scherer.
\newblock Voxnet: A 3d convolutional neural network for real-time object
  recognition.
\newblock In {\em 2015 IEEE/RSJ International Conference on Intelligent Robots
  and Systems (IROS)}, pages 922--928. IEEE, 2015.

\bibitem{mescheder2019occupancy}
Lars Mescheder, Michael Oechsle, Michael Niemeyer, Sebastian Nowozin, and
  Andreas Geiger.
\newblock Occupancy networks: Learning 3d reconstruction in function space.
\newblock In {\em Proceedings of the IEEE/CVF Conference on Computer Vision and
  Pattern Recognition}, pages 4460--4470, 2019.

\bibitem{mildenhall2020nerf}
Ben Mildenhall, Pratul~P Srinivasan, Matthew Tancik, Jonathan~T Barron, Ravi
  Ramamoorthi, and Ren Ng.
\newblock Nerf: Representing scenes as neural radiance fields for view
  synthesis.
\newblock In {\em European conference on computer vision}, pages 405--421.
  Springer, 2020.

\bibitem{mo2019partnet}
Kaichun Mo, Shilin Zhu, Angel~X Chang, Li Yi, Subarna Tripathi, Leonidas~J
  Guibas, and Hao Su.
\newblock Partnet: A large-scale benchmark for fine-grained and hierarchical
  part-level 3d object understanding.
\newblock In {\em Proceedings of the IEEE/CVF Conference on Computer Vision and
  Pattern Recognition}, pages 909--918, 2019.

\bibitem{NimierDavidVicini2019Mitsuba2}
Merlin Nimier-David, Delio Vicini, Tizian Zeltner, and Wenzel Jakob.
\newblock Mitsuba 2: A retargetable forward and inverse renderer.
\newblock {\em Transactions on Graphics (Proceedings of SIGGRAPH Asia)}, 38(6),
  Dec. 2019.

\bibitem{novotny2019c3dpo}
David Novotny, Nikhila Ravi, Benjamin Graham, Natalia Neverova, and Andrea
  Vedaldi.
\newblock C3dpo: Canonical 3d pose networks for non-rigid structure from
  motion.
\newblock In {\em Proceedings of the IEEE/CVF International Conference on
  Computer Vision}, pages 7688--7697, 2019.

\bibitem{palmer1999vision}
Stephen~E Palmer.
\newblock {\em Vision science: Photons to phenomenology}.
\newblock MIT press, 1999.

\bibitem{park2019deepsdf}
Jeong~Joon Park, Peter Florence, Julian Straub, Richard Newcombe, and Steven
  Lovegrove.
\newblock Deepsdf: Learning continuous signed distance functions for shape
  representation.
\newblock In {\em Proceedings of the IEEE/CVF Conference on Computer Vision and
  Pattern Recognition}, pages 165--174, 2019.

\bibitem{poulenard2021functional}
Adrien Poulenard and Leonidas~J Guibas.
\newblock A functional approach to rotation equivariant non-linearities for
  tensor field networks.
\newblock In {\em Proceedings of the IEEE/CVF Conference on Computer Vision and
  Pattern Recognition}, pages 13174--13183, 2021.

\bibitem{qi2017pointnet}
Charles~R Qi, Hao Su, Kaichun Mo, and Leonidas~J Guibas.
\newblock Pointnet: Deep learning on point sets for 3d classification and
  segmentation.
\newblock In {\em Proceedings of the IEEE conference on computer vision and
  pattern recognition}, pages 652--660, 2017.

\bibitem{qi2016volumetric}
Charles~R Qi, Hao Su, Matthias Nie{\ss}ner, Angela Dai, Mengyuan Yan, and
  Leonidas~J Guibas.
\newblock Volumetric and multi-view cnns for object classification on 3d data.
\newblock In {\em Proceedings of the IEEE conference on computer vision and
  pattern recognition}, pages 5648--5656, 2016.

\bibitem{qi2017pointnetpp}
Charles~Ruizhongtai Qi, Li Yi, Hao Su, and Leonidas~J Guibas.
\newblock Pointnet++: Deep hierarchical feature learning on point sets in a
  metric space.
\newblock In {\em NIPS}, pages 5099--5108, 2017.

\bibitem{reizenstein2021common}
Jeremy Reizenstein, Roman Shapovalov, Philipp Henzler, Luca Sbordone, Patrick
  Labatut, and David Novotny.
\newblock Common objects in 3d: Large-scale learning and evaluation of
  real-life 3d category reconstruction.
\newblock In {\em Proceedings of the IEEE/CVF International Conference on
  Computer Vision}, pages 10901--10911, 2021.

\bibitem{rusu2009fast}
Radu~Bogdan Rusu, Nico Blodow, and Michael Beetz.
\newblock Fast point feature histograms (fpfh) for 3d registration.
\newblock In {\em 2009 IEEE international conference on robotics and
  automation}, pages 3212--3217. IEEE, 2009.

\bibitem{sajnani2021draco}
Rahul Sajnani, AadilMehdi Sanchawala, Krishna~Murthy Jatavallabhula, Srinath
  Sridhar, and K~Madhava Krishna.
\newblock Draco: Weakly supervised dense reconstruction and canonicalization of
  objects.
\newblock In {\em 2021 IEEE International Conference on Robotics and Automation
  (ICRA)}, pages 10302--10309. IEEE, 2021.

\bibitem{salti2014shot}
Samuele Salti, Federico Tombari, and Luigi Di~Stefano.
\newblock Shot: Unique signatures of histograms for surface and texture
  description.
\newblock {\em Computer Vision and Image Understanding}, 125:251--264, 2014.

\bibitem{shepard1971mental}
Roger~N Shepard and Jacqueline Metzler.
\newblock Mental rotation of three-dimensional objects.
\newblock {\em Science}, 171(3972):701--703, 1971.

\bibitem{shotton2011real}
Jamie Shotton, Andrew Fitzgibbon, Mat Cook, Toby Sharp, Mark Finocchio, Richard
  Moore, Alex Kipman, and Andrew Blake.
\newblock Real-time human pose recognition in parts from single depth images.
\newblock In {\em CVPR 2011}, pages 1297--1304. Ieee, 2011.

\bibitem{shotton2013scene}
Jamie Shotton, Ben Glocker, Christopher Zach, Shahram Izadi, Antonio Criminisi,
  and Andrew Fitzgibbon.
\newblock Scene coordinate regression forests for camera relocalization in
  rgb-d images.
\newblock In {\em Proceedings of the IEEE Conference on Computer Vision and
  Pattern Recognition}, pages 2930--2937, 2013.

\bibitem{spezialetti2020learning}
Riccardo Spezialetti, Federico Stella, Marlon Marcon, Luciano Silva, Samuele
  Salti, and Luigi Di~Stefano.
\newblock Learning to orient surfaces by self-supervised spherical cnns.
\newblock {\em arXiv preprint arXiv:2011.03298}, 2020.

\bibitem{sridhar2019multiview}
Srinath Sridhar, Davis Rempe, Julien Valentin, Sofien Bouaziz, and Leonidas~J
  Guibas.
\newblock Multiview aggregation for learning category-specific shape
  reconstruction.
\newblock {\em arXiv preprint arXiv:1907.01085}, 2019.

\bibitem{su2015multi}
Hang Su, Subhransu Maji, Evangelos Kalogerakis, and Erik Learned-Miller.
\newblock Multi-view convolutional neural networks for 3d shape recognition.
\newblock In {\em Proc. ICCV}, pages 945--953, 2015.

\bibitem{sun2020canonical}
Weiwei Sun, Andrea Tagliasacchi, Boyang Deng, Sara Sabour, Soroosh Yazdani,
  Geoffrey Hinton, and Kwang~Moo Yi.
\newblock Canonical capsules: Unsupervised capsules in canonical pose.
\newblock {\em arXiv preprint arXiv:2012.04718}, 2020.

\bibitem{tarr1989mental}
Michael~J Tarr and Steven Pinker.
\newblock Mental rotation and orientation-dependence in shape recognition.
\newblock {\em Cognitive psychology}, 21(2):233--282, 1989.

\bibitem{tatarchenko2019single}
Maxim Tatarchenko, Stephan~R Richter, Ren{\'e} Ranftl, Zhuwen Li, Vladlen
  Koltun, and Thomas Brox.
\newblock What do single-view 3d reconstruction networks learn?
\newblock In {\em Proceedings of the IEEE/CVF Conference on Computer Vision and
  Pattern Recognition}, pages 3405--3414, 2019.

\bibitem{thomas2018tensor}
Nathaniel Thomas, Tess Smidt, Steven Kearnes, Lusann Yang, Li Li, Kai Kohlhoff,
  and Patrick Riley.
\newblock Tensor field networks: Rotation-and translation-equivariant neural
  networks for 3d point clouds.
\newblock {\em arXiv preprint arXiv:1802.08219}, 2018.

\bibitem{wang20206}
Chen Wang, Roberto Mart{\'\i}n-Mart{\'\i}n, Danfei Xu, Jun Lv, Cewu Lu, Li
  Fei-Fei, Silvio Savarese, and Yuke Zhu.
\newblock 6-pack: Category-level 6d pose tracker with anchor-based keypoints.
\newblock In {\em 2020 IEEE International Conference on Robotics and Automation
  (ICRA)}, pages 10059--10066. IEEE, 2020.

\bibitem{wang2015voting}
Dominic~Zeng Wang and Ingmar Posner.
\newblock Voting for voting in online point cloud object detection.
\newblock In {\em Robotics: Science and Systems}, volume~1, pages 10--15. Rome,
  Italy, 2015.

\bibitem{wang2019normalized}
He Wang, Srinath Sridhar, Jingwei Huang, Julien Valentin, Shuran Song, and
  Leonidas~J Guibas.
\newblock Normalized object coordinate space for category-level 6d object pose
  and size estimation.
\newblock In {\em Proceedings of the IEEE/CVF Conference on Computer Vision and
  Pattern Recognition}, pages 2642--2651, 2019.

\bibitem{wang2019deep}
Yue Wang and Justin~M Solomon.
\newblock Deep closest point: Learning representations for point cloud
  registration.
\newblock In {\em Proceedings of the IEEE/CVF International Conference on
  Computer Vision}, pages 3523--3532, 2019.

\bibitem{wang2018dynamic}
Yue Wang, Yongbin Sun, Ziwei Liu, Sanjay~E Sarma, Michael~M Bronstein, and
  Justin~M Solomon.
\newblock Dynamic graph cnn for learning on point clouds.
\newblock {\em arXiv preprint arXiv:1801.07829}, 2018.

\bibitem{weiler20183d}
Maurice Weiler, Mario Geiger, Max Welling, Wouter Boomsma, and Taco Cohen.
\newblock 3d steerable cnns: Learning rotationally equivariant features in
  volumetric data.
\newblock In {\em NIPS}, pages 10381--10392, 2018.

\bibitem{wu20153d}
Zhirong Wu, Shuran Song, Aditya Khosla, Fisher Yu, Linguang Zhang, Xiaoou Tang,
  and Jianxiong Xiao.
\newblock 3d shapenets: A deep representation for volumetric shapes.
\newblock In {\em Proc. CVPR}, pages 1912--1920, 2015.

\bibitem{yang2019pointflow}
Guandao Yang, Xun Huang, Zekun Hao, Ming-Yu Liu, Serge Belongie, and Bharath
  Hariharan.
\newblock Pointflow: 3d point cloud generation with continuous normalizing
  flows.
\newblock In {\em Proceedings of the IEEE/CVF International Conference on
  Computer Vision}, pages 4541--4550, 2019.

\bibitem{yang2021continuous}
Zhangsihao Yang, Or Litany, Tolga Birdal, Srinath Sridhar, and Leonidas Guibas.
\newblock Continuous geodesic convolutions for learning on 3d shapes.
\newblock In {\em Proceedings of the IEEE/CVF Winter Conference on Applications
  of Computer Vision}, pages 134--144, 2021.

\bibitem{deepgmr}
Wentao Yuan, Benjamin Eckart, Kihwan Kim, Varun Jampani, Dieter Fox, and Jan
  Kautz.
\newblock Deepgmr: Learning latent gaussian mixture models for registration.
\newblock {\em CoRR}, abs/2008.09088, 2020.

\bibitem{yuan2020deepgmr}
Wentao Yuan, Benjamin Eckart, Kihwan Kim, Varun Jampani, Dieter Fox, and Jan
  Kautz.
\newblock Deepgmr: Learning latent gaussian mixture models for registration.
\newblock In {\em European Conference on Computer Vision}, pages 733--750.
  Springer, 2020.

\bibitem{zhang2018learning}
Xiuming Zhang, Zhoutong Zhang, Chengkai Zhang, Joshua~B Tenenbaum, William~T
  Freeman, and Jiajun Wu.
\newblock Learning to reconstruct shapes from unseen classes.
\newblock {\em arXiv preprint arXiv:1812.11166}, 2018.

\bibitem{zhang2019rotation}
Zhiyuan Zhang, Binh-Son Hua, David~W Rosen, and Sai-Kit Yeung.
\newblock Rotation invariant convolutions for 3d point clouds deep learning.
\newblock In {\em 2019 International Conference on 3D Vision (3DV)}, pages
  204--213. IEEE, 2019.

\end{thebibliography}
}
\hfill
\vfill

\newpage
\appendix

\begin{center}{\bf {\LARGE 
Appendix \\ [1em]
}
}
\end{center}
\section{Network details}

\subsection{Architecture}

We reuse the classification architecture described in Section 3.1 of \cite{poulenard2021functional} as our backbone. The architecture comprises of three equivariant convolution layers followed by a global max-pooling layer, and the remaining layers specialize for classification; we drop these last layers and specialize the network for our tasks instead. The global max-pooling layer of \cite{poulenard2021functional} proceeds by first interpreting each point-wise signal as coefficients of spherical functions in the SH basis and performing a discrete inverse spherical harmonics transform to convert them into functions over a discrete sampling of the sphere. For any direction, the resulting signal is then spatially pooled over the shape, resulting in a single function over the sphere sampling (specifically, a single map from the sphere sampling to $\mathbf{R}^C$, where $C = 256$ as we have 256 channels). We then apply point-wise MLPs (with ReLU activations) on this sphere map and convert it back to TFN-like features via forward spherical harmonics transform (SHT) \cite{poulenard2021functional}. 

    \parahead{Spherical Harmonic Coefficients} In order to predict the coefficients $F(X)$ of the invariant embedding $H(X)$, we apply a $[128, 64]$-MLP whose last layer is linear and convert to types $\ell \in \llbracket 0, 3\rrbracket$ via SHT.
    
    \parahead{Rotation-Invariant Point Cloud} We obtain our 3D invariant point cloud $X^c$ by applying a linear layer to $H(X)$.
    
    \parahead{Rotation-Equivariant Frame} To predict $E$, we apply a $[64, 3]$-MLP whose last layer has a linear activation. We then extract type $1$ features with SHT, giving us a collection of 3 equivariant 3D vectors.
    
    \parahead{Segmentation} To predict the segmentation we apply a point-wise $[256, 128, 10]$-MLP whose last layer is soft-max to get the segmentation masks $S$ described in \cref{sec:segmentation}.

\subsection{Training Details}

\parahead{Cropping operator $\mathcal{O}$} We introduce synthetic occlusion in our training setting by slicing full shapes using the cropping operator $\mathcal{O}$. To perform a crop, we uniformly sample a direction $v$ on the unit sphere and remove the top $K/2$ points in the shape that have the highest value of $x^Tv$ for $x \in X$. Additionally, we train our model on the ShapeNetCOCO dataset~\cite{sridhar2019multiview,sajnani2021draco} which has pre-determined occlusion due to camera motion, as seen in~\cref{fig:nocs}. In order to preprocess this data for training, we aggregate the parts in the canonical NOCS space of every sequence to obtain the full shape and perform a nearest neighbor search in the NOCS space to find correspondences between the full and partial shape.

\parahead{Hyper-parameters} During training, we use a batch size of $16$ in every step for all our models. We set an $L^1$ kernel regularizer at every layer of the network with weight $0.1$. We weigh the loss functions by their effect on reducing the Canonical Shape loss $\mathcal{L}_{canon}$. The loss functions are weighed as: $\mathcal{L}_{canon} \;(2)$, $\mathcal{L}_{rest}\; (1)$, $\mathcal{L}_{ortho}\; (1)$, $\mathcal{L}_{sep}\; (0.8)$, and $\mathcal{L}_{amod}\; (1)$.

\section{Unsupervised Co-segmentation}
\label{sec:segmentation}
\subsection{Predicting parts} We predict the part segments $S \in \mathbb{R}^{K \times C}$ wherein $C$ are the number of parts. We use the rotation-invariant embedding $H^{\ell}(X)$ with all the types $0 \leq \ell \leq 3$ to predict the segmentation $S$. We define the following notation for normalized parts $A(X)$ and part centroids $\theta(X)$ similar to \cite{sun2020canonical}.:
\begin{equation}
    \begin{split}
        & S(X) := \mathrm{Softmax}[\mathrm{MLP}(H(X))] \\
        & \; A_{ij}(X) := \frac{S_{ij}(X)}{\sum_i S_{ij}(X)} \\ 
        & \theta_{j}(X) := \sum_i A_{ij}(X)X_{i,:}
    \end{split}
\end{equation}

\subsection{Loss functions}
We use part segmentation to enforce semantic consistency between full and partial shapes. We borrow the localization loss ($\mathcal{L}_{localization}$) and equilibrium loss ($\mathcal{L}_{equilibrium}$) from \cite{sun2020canonical} for the full shape to evenly spread part segmentation across the shape. Additionally, we employ the following losses.

\parahead{Part Distribution loss} We compute the two-way Chamfer distance ($CD$) between the part centroids and the input shape. In practice, this helps to distribute parts more evenly across the shape.
\begin{equation}
    \mathcal{L}_{dist} = \mathrm{CD}\left( X, \theta(X) \right)
        \label{eq:capsule_chamfer_loss}
\end{equation}

\parahead{Part Restriction loss} The parts discovered by the network for the partial shape should be congruent to the parts discovered by the network for the full shape. We penalize the part prediction for corresponding parts by minimizing the negative Cosine Similarity ($CS$) for our capsule predictions.
\begin{equation}
    \mathcal{L}_{rest(part)} =  - \frac{2}{K}\sum_{i \in \mathcal{S}} \mathbf{CS}(S(\mathcal{O}(X))_{i,:}, \mathcal{O}(S(X))_{i,:}) 
\end{equation}

\parahead{Part Directional loss} To avoid part centers of the visible parts of a shape from deviating from the part centers of the full shape, we use a soft loss to ensure that the directional vector between part centers are consistent between the full and partial shape. $\mathrm{dir}(\theta(X))$ computes the vector directions between every $^{C}C_2$ centroid pairs for $C$ part centroids. 
\begin{equation}
    \mathcal{L}_{direc} = - \frac{1}{^{C}C_2} \sum_{i \in \mathcal{S}} \mathbf{CS}(\mathrm{dir}(\theta(\mathcal{O}(X_i))), \mathrm{dir}(\mathcal{O}(\theta(X_i))))
\end{equation}


\section{Registration}
\label{sec:registration}

We note in \cref{table:registration_shapenet} that our method does not perform well in this task as we predict a frame $E \in O(3)$ which can have reflection symmetries, we observe symmetries such as left-right reflection for planes. Symmetries cause high RMSE error because points are matched with their image under symmetry which are often very distant. However, when using Chamfer Distance metric which is symmetry agnostic our registration error decreases by an order of magnitude achieving competitive results on this benchmark. We also note that Ours(F+P) noticeably decreases $\mathrm{RMSE}$ compared to  Ours(F) as during training the frame consistency is enforced between the full shape and a randomly rotated partial by the $\mathcal{L}_{rest}$ loss.



\section{Ablations}
We now provide detailed ablations to justify the following key design choices: the effect of increasing amounts of occlusion/partiality, and loss
functions.

\parahead{Degree of Occlusion/Partiality} We examine the ability of our model to handle varying amounts of occlusion/partiality for the car category in Table~\ref{table:deg_partiality}.
Our occlusion function, $\mathcal{O}$, occludes shapes to only keep a fraction of the original shape between 25\% and 75\% (\ie,~75\% is more occluded than 25\%).
We observe that our method performs optimally over all metrics when trained at 50\% occlusion.
%
\vspace{1cm}

\begin{table}[h]
\centering
\scalebox{0.85}{
\begin{tabular}{c|cccc}
\toprule
\multicolumn{1}{c|}{{\textbf{Test partiality}}}                          & \multicolumn{4}{c}{\textbf{Degree of partiality during training}}                                  \\ \cmidrule{2-5} 
\multicolumn{1}{c|}{}                                                          & \multicolumn{1}{c}{75\%} & \multicolumn{1}{c}{50\%} & \multicolumn{1}{c}{25\%} & {[}25\%, 75\%{]}  \\ \midrule
\multicolumn{5}{l}{\textbf{Ground Truth Consistency (GC)}{\color{blue} $\downarrow$}} \\
\midrule
\multicolumn{1}{l|}{75\%}            & 0.0451                   & \textbf{0.0438}          & 0.1420                   & 0.0681          \\ 
\multicolumn{1}{l|}{50\%}            & 0.0375                   & 0.0356                   & 0.0504                   & \textbf{0.0296} \\ 
\multicolumn{1}{l|}{25\%}            & 0.0388                   & 0.0301                   & \textbf{0.0241}          & 0.0299          \\ 
\multicolumn{1}{l|}{{[}25\%, 75\%{]}} & \textbf{0.0438}          & 0.0553                   & 0.0894                   & 0.0558          \\ \midrule
\multicolumn{5}{l}{\textbf{Instance-Level Consistency (IC)}{\color{blue} $\downarrow$}} \\
\midrule
\multicolumn{1}{l|}{75\%}            & 0.0728                   & \textbf{0.0719}          & 0.1542                   & 0.0797          \\ 
\multicolumn{1}{l|}{50\%}            & 0.0452                   & \textbf{0.0349}          & 0.0526                   & 0.0380          \\ 
\multicolumn{1}{l|}{25\%}            & 0.0456                   & 0.0333                   & \textbf{0.0221}          & 0.0334          \\ 
\multicolumn{1}{l|}{{[}25\%, 75\%{]}} & \textbf{0.0719}          & 0.0792                   & 0.1049                   & 0.0804          \\ \midrule
\multicolumn{5}{l}{\textbf{Category-Level Consistency (CC)}{\color{blue} $\downarrow$}} \\
\midrule
\multicolumn{1}{l|}{75\%}            & 0.0914                   & \textbf{0.0895}          & 0.1702                   & 0.0966          \\ 
\multicolumn{1}{l|}{50\%}            & 0.0652                   & 0.0632                   & 0.0731                   & \textbf{0.0617} \\ 
\multicolumn{1}{l|}{25\%}            & 0.0657                   & 0.0608                   & \textbf{0.0582}          & 0.0606          \\ 
\multicolumn{1}{l|}{{[}25\%, 75\%{]}} & \textbf{0.0895}          & 0.0985                   & 0.1216                   & 0.0982          \\ \midrule
\multicolumn{1}{l|}{\textbf{Average}}                                          & 0.0594                   & \textbf{0.0580}          & 0.0886                   & 0.0610          \\ \midrule
\end{tabular}
}
\caption{\textbf{Degree of partiality} - Partiality introduced during training (vertical) is evaluated on the canonicalization metrics with different fraction of partiality (horizontal). [25\%, 75\%] indicates that degrees of partiality between 25\% and 75\%  are randomly introduced in the shapes. Our model trained with partiality $50\%$ performs better on average over all the canonicalization metrics. [\textit{Note: 75\%  is  more  occluded  than 25\%. }] \label{table:deg_partiality}} 
\end{table}

\begin{table*}[!hbt]
\small
\centering
\scalebox{0.9}{
\begin{tabular}{c|ccc|ccc|ccc|ccc}
\toprule
\textbf{Category}  $\rightarrow$     & \multicolumn{3}{c|}{\textbf{Plane}}                                                                    & \multicolumn{3}{c|}{\textbf{Table}}                                                                        & \multicolumn{3}{c|}{\textbf{Chair}}  & \multicolumn{3}{c}{\textbf{Average}}  \\ \midrule
\textbf{Metric} $\downarrow$        & \textbf{Ours} & \textbf{w/o sep} & \textbf{w/o rest}      & \textbf{Ours} & \textbf{w/o sep}      & \textbf{w/o rest}      & \textbf{Ours} & \textbf{w/o sep}      & \textbf{w/o rest} & \textbf{Ours} & \textbf{w/o sep}      & \textbf{w/o rest}  \\ \midrule
\multicolumn{1}{l|}{GC (full)}     &\textbf{0.0286}         &0.0321        &0.0303               &0.0738                  &\textbf{0.0641}          &0.0729     &0.0509           &\textbf{0.0430}              &0.0532 & 0.0511    & \textbf{0.0464} & 0.0521           \\ 

\multicolumn{1}{l|}{IC (full)}      &\textbf{0.0144}         &0.0187        &0.0169               &\textbf{0.0361}         &0.0612                   &0.0411     &0.0235           &\textbf{0.0224}              &0.0245    & \textbf{0.0247} & 0.0341 & 0.0275          \\ 

\multicolumn{1}{l|}{CC (full)}    &\textbf{0.0679}         &0.0697        &0.0683               &\textbf{0.1432}         &0.1510                   &0.1434     &0.1145           &0.1150              &\textbf{0.1143}    & \textbf{0.1085} & 0.1119 & 0.1087  \\ 
\midrule
\multicolumn{1}{l|}{GC (partial)}  &0.0360                  &0.0389        &\textbf{0.0332}      &0.0662                  &\textbf{0.0523}          &0.0683     &0.0780           &\textbf{0.0681}              &0.0850 & 0.0601 & \textbf{0.0531} & 0.0622              \\ 
\multicolumn{1}{l|}{IC (partial)}   &\textbf{0.0265}         &0.0324        &0.0479               &\textbf{0.0739}         &0.0791                   &0.0805     &0.0622           &\textbf{0.0537}              &0.0841              & \textbf{0.0542} & 0.0551 & 0.0708 \\ 

\multicolumn{1}{l|}{CC (partial)} &\textbf{0.0713}         &0.0733        &0.0765               &\textbf{0.1579}         &0.1590                   &0.1598     &0.1270           &\textbf{0.1250}              &0.1377    & \textbf{0.1187} & 0.1191 & 0.1247            \\ \midrule
\multicolumn{1}{l|}{Average} & \textbf{0.0408} & 0.0442 & 0.0455 & \textbf{0.0912} & 0.0945 & 0.0943 & 0.0760 & \textbf{0.0712} & 0.0831 & \textbf{0.0696} & 0.0700 & 0.0743
\\
\bottomrule
\end{tabular}
}
\caption{Ablation study to investigate the
effect of different loss functions. "w/o sep" and "w/o rest" denote training without separation and without restriction loss, respectively.\label{table_supp:loss_ablation}}
\end{table*}

\parahead{Loss Functions}
We evaluate our F+P model on both full
and partial shapes trained
with all losses, without the separation loss $\mathcal{L}_{sep}$, and without the restriction loss $\mathcal{L}_{rest}$. From Table~\ref{table_supp:loss_ablation}, we observe that using restriction loss $\mathcal{L}_{rest}$ helps in canonicalization of both full and partial shapes in categories \textit{plane}, \textit{table}, and \textit{chair}. However, separation loss, $\mathcal{L}_{sep}$, helps in \textit{plane}, \textit{table} but not in \textit{chair}. Since, both losses help in most of the categories, we utilize them for training our final model.

\parahead{Effect of introducing occlusion on full shapes}
We evaluate the canonicalization of full shapes using our network trained on full and partial shapes. We observe that on average both our models \textbf{Ours(F)} and \textbf{Ours(F+P)} perform the same on the canonicalization metrics for full shapes. For a few categories such as \textit{lamp, car, chair, watercraft}, introducing partial shapes in the training improves its performance on the canonicalization metrics. Whereas introducing occlusion during training degrades the performance for category $bench$.

\section{Applications}

\subsection{Co-Canonicalization}
\begin{figure}[!hbt]
\centering
  \includegraphics[width=0.5\textwidth]{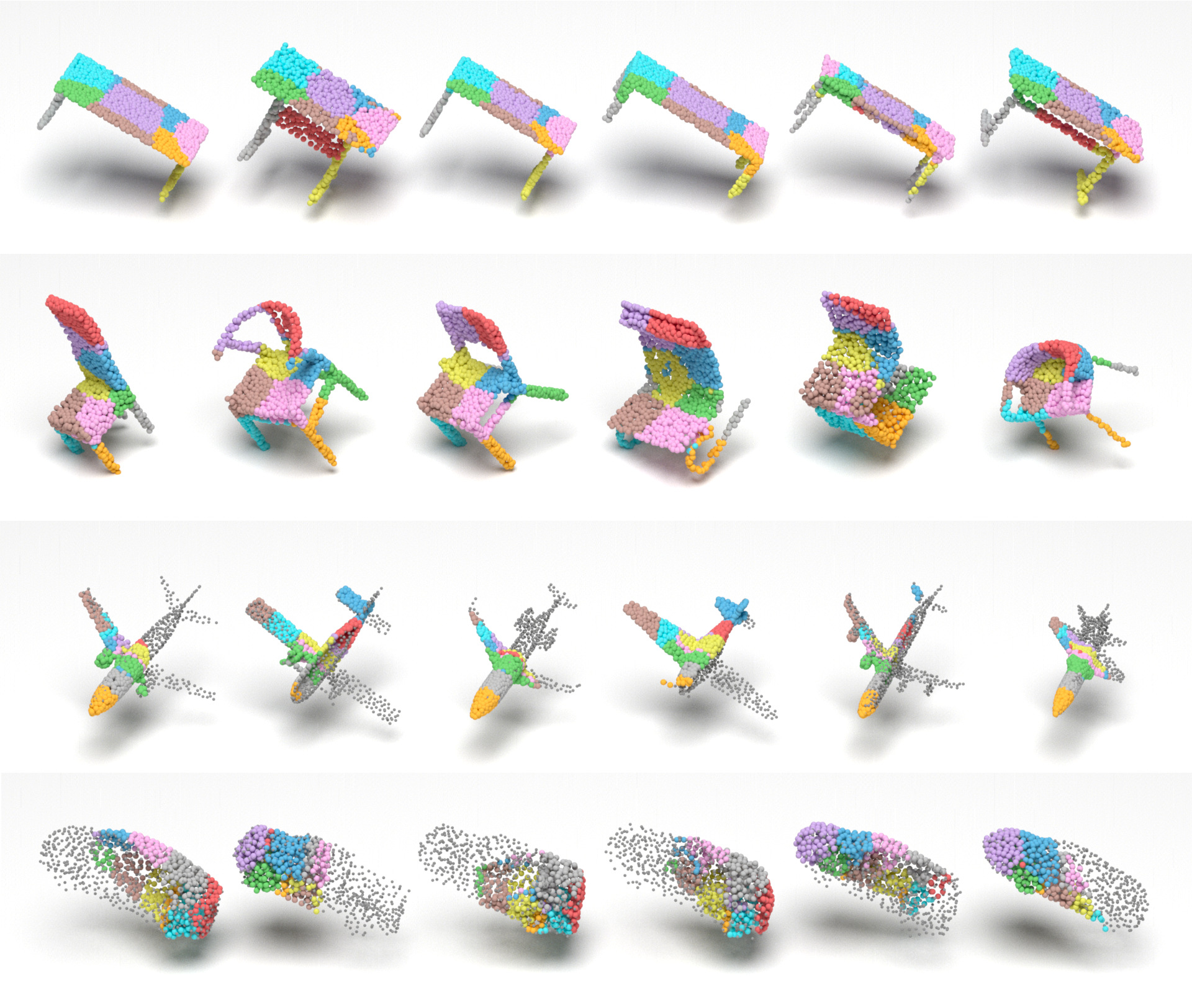}
  \caption{Co-canonicalizing object instances from ModelNet40 using our method trained on ShapeNet (core). (\textbf{\emph{top}}) Canonicalized full shapes. (\textbf{\emph{bottom}}) Canonicalized partial shapes.}
  \label{fig:single_cocan}
\end{figure}

Commonly used datasets in 3D vision, such as ShapeNet~\cite{chang2015shapenet}, are manually pre-canonicalized, making expansion of such datasets expensive. Since our method performs better than others on canonicalization, we believe that it can be used to extend these datasets by canonicalizing corpora of \textit{in-the-wild} shapes into a common pose. Figure~\ref{fig:single_cocan} shows the results of our model, trained on the ShapeNet (core) dataset~\cite{chang2015shapenet}, being used to canonicalize shapes from the (uncanonicalized) ModelNet40 dataset~\cite{wu20153d}. These shapes can now be merged into ShapeNet by applying a single category-wide rotation to match the obtained canonical frame with the existing frame used by ShapeNet, instead of the per-instance rotation that would otherwise be required. Furthermore, these results qualitatively demonstrate the ability of our method to generalize to datasets not seen during training.


\subsection{Depth Map Canonicalization}
\begin{figure}[!hbt]
\centering
  \includegraphics[width=0.45\textwidth]{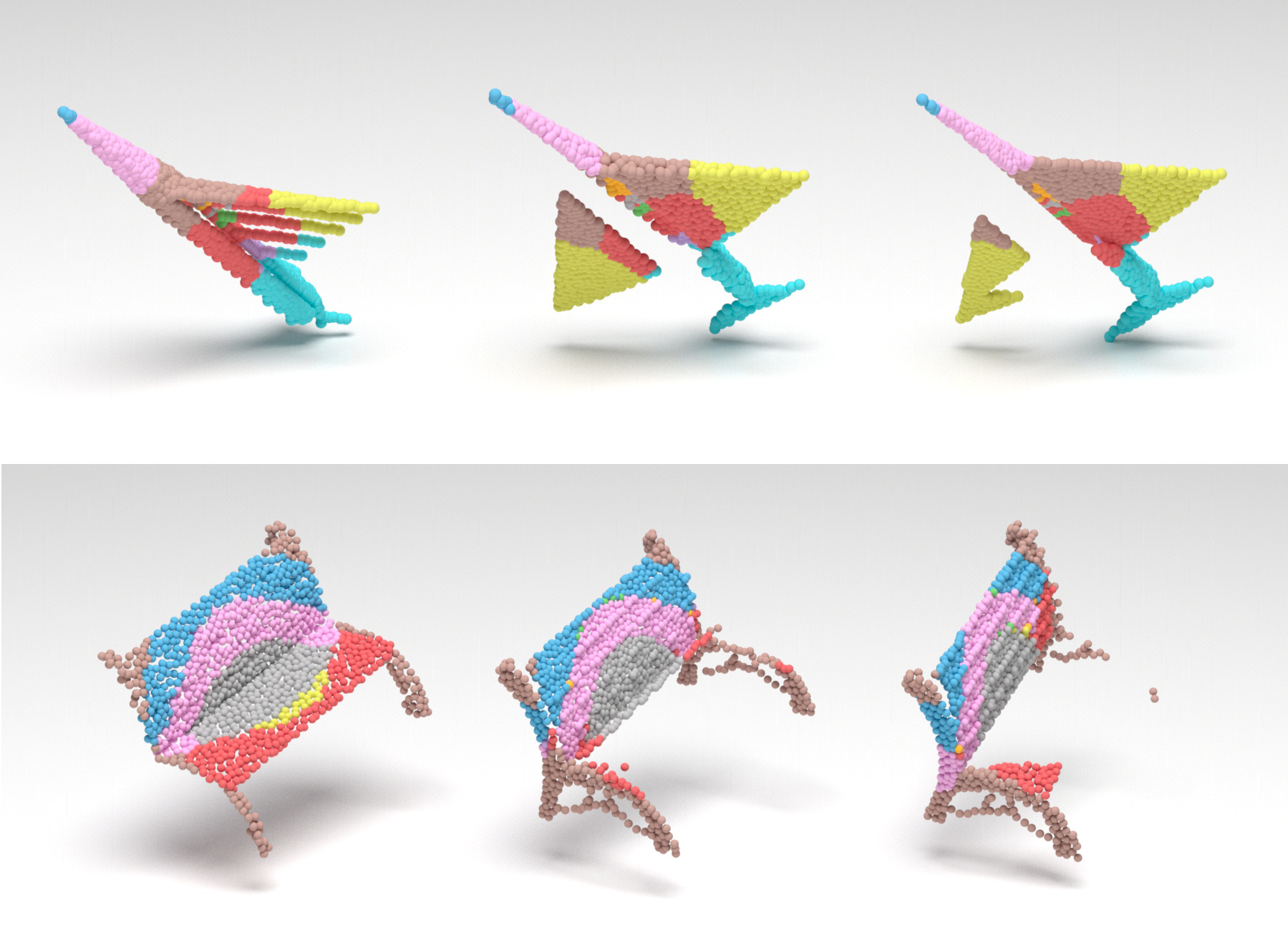}
  \caption{Canonicalizing point clouds obtained from depth maps from the ShapeNetCOCO dataset.}
  \label{fig:nocs}
\end{figure}

Since our method operates on partial shapes, we can canonicalize objects in \textbf{depth images}. We use the depth maps from the ShapeNetCOCO dataset, which have pre-determined occlusion due to camera motion, and canonicalize partial point clouds.
Specifically, we first take depth maps and utilize them to generate groundtruth pointclouds. We then trained and tested our model on it. Figure~\ref{fig:nocs} present examples to demonstrate that our model is capable of canonicalizing depth maps.

\subsection{Annotation Transfer}
\begin{figure}[!h]
\centering
  \includegraphics[width=0.45\textwidth]{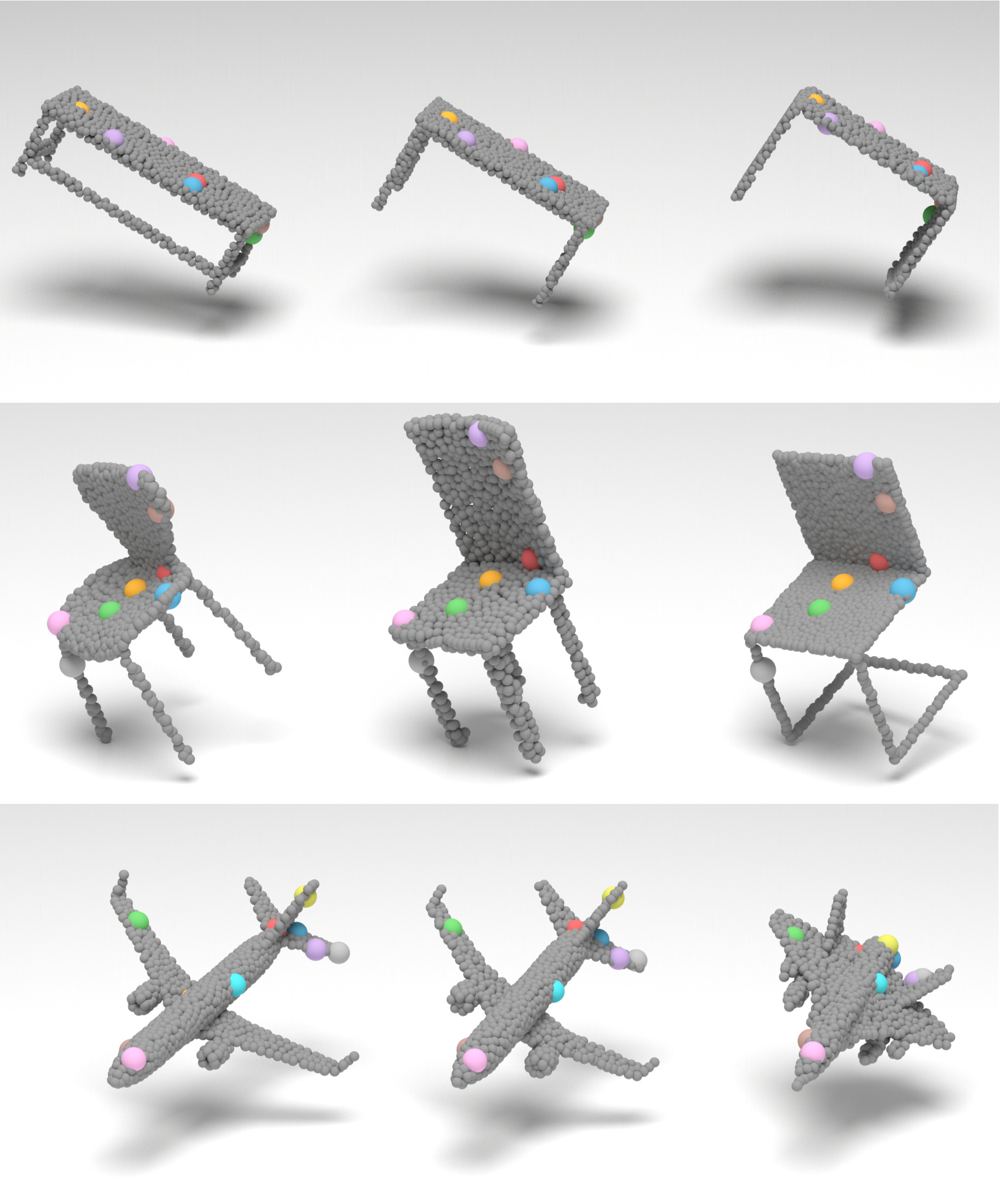}
  \caption{Transferring key-point annotations from one shape to another in the same category. We annotate only the first column of shapes and transfer key-points to all the other columns}
  \label{fig:single_annotation}
\end{figure}

Since a category-level canonical frame is consistent with respect to the geometry and local shape of different object instances of a category, annotations can be transferred across instances that share the same canonical frame. Particularly, we demonstrate the transfer of sparse key-point annotations in Figure~\ref{fig:single_annotation}. We randomly assign labels to a few points of one point cloud in each category, which serves as the source. We then use a remarkably simple transfer function to transfer these labels to points in each target point cloud, making use of the predicted segmentation. To every labeled point in the source point cloud, we obtain a directional vector originating from the centroid of the segment it belongs to. Starting from the corresponding centroid in the target point cloud, we move along this directional vector and then pick the nearest point. While this scheme works well in our case, more nuanced transfer functions may be required depending on the application.

\section{Proof of Rotation-Invariance Property of our Embedding}

Given rotation-equivariant embeddings $F^{\ell}$ and $Y^{\ell}$ the tensors $H^{\ell}(X)$ are rotation invariant as:
\[
\begin{aligned}
H^{\ell}_{ijk}(R.X)
&
= 
\langle F^{\ell}_{i,:,j}(R.X), Y^{\ell}_{:,j,k}(R.X)\rangle
\\
&
=
\langle D^{\ell}(R)F^{\ell}_{i,:,j}(X), D^{\ell}(R) Y^{\ell}_{:,j,k}(X)\rangle
\\
&
=
\langle F^{\ell}_{i,:,j}(X),  Y^{\ell}_{:,j,k}(X)\rangle
=
H^{\ell}_{ijk}(X)
\end{aligned}
\]

\section{Commutative Property of Canonicalization with the Cropping Operator}
Canonicalization commutes with the cropping operator $\mathcal{O}$. For a (full) point cloud $X$ and predicted canonicalizing frame $\mathcal{R}(X)$, we prove the commutative property here, we assume $X$ is mean centered for simplification.
\[
\begin{aligned}
    &\widehat{\mathcal{O}[X]}^c + \mathcal{R}(X) \overline{\mathcal{O}[X]}
    = \mathcal{R}(X)(\widehat{\mathcal{O}[X]} + \overline{\mathcal{O}[X]})
    \\ 
    &=
    \mathcal{R}(X)(\mathcal{O}[X]) =\mathcal{O}[\mathcal{R}(X)X]
    =\mathcal{O}[X^c]
\end{aligned}
\]
The above commutative property enables us to a predict a rotation-equivariant translation $\mathcal{T}(\widehat{\mathcal{O}(X)})$ from the mean centered partial shape $\widehat{\mathcal{O}(X)}$ only that aligns the partial shape to its corresponding points in the full shape.
\[
\begin{aligned}
    &\widehat{\mathcal{O}[X]}^c + \mathcal{R}(\widehat{\mathcal{O}[X]}) \overline{\mathcal{O}[X]} \simeq \widehat{\mathcal{O}[X]}^c + \mathcal{R}(\widehat{\mathcal{O}[X]}) \mathcal{T}(\widehat{\mathcal{O}(X)}) 
    \\
    &=\mathcal{O}[X^{c}] 
\end{aligned}
\]


\section{Discussion on Canonicalization Metrics}

We complement the discussion of our canonicalization metrics with a few remarks. Our 3 metrics Instance-Level (IC), Category-Level (CC) and Ground Truth (GC) Consistency measure three aspects of canonicalization. The instance-level metric is a measure of the "variance" of the canonical pose under rotation of the input. By definition the canonical pose must be invariant to the input pose. The GC metric provides a way of measuring canonicalization consistency across the entire class of objects by measuring how our canonicalization deviates from a ground truth canonicalization up to a constant rotation. In the absence of a ground truth alignment, we propose the CC metric which compares canonicalization of different shapes within the same class using Chamfer distance (as we don't assume pointwise correspondences between different shapes). The CC metric relies on the assumption that aligned shapes of the same category are similar to each other.

We observe in table (1) of our article that some methods have high IC but low GC and vice versa (e.g. CaCa \cite{sun2020canonical} (cabinet), Ours (F + P) speaker). This occurs as we canonicalize based on geometric similarity instead of semantic aspects of the object. The IC and CC metrics measure geometric properties of the canonicalization while GC measures semantic properties of the canonicalization according to manually aligned shapes.

We build our metrics using the Chamfer distance as it does not assume pointwise correspondences between shapes, this allows measuring the canonicalization quality of symmetric shapes where there may not be a single correct canonical orientation. However, we observe a performance gap with our method when using distances based on pointwise correspondences such as $L^2$ or root mean square (RMSE) errors as seen in \cref{sec:registration} of this appendix. We believe our Chamfer distance based metrics are representative of the quality of canonicalization and are consistent with our visual evaluation.

\section{Discussion on PCA}
\parahead{PCA Over-Performance on the CC Metric}
We note that the competitiveness of PCA is limited to certain experiments for \textbf{full shapes} and \textbf{multi-category} experiments only. The CC metric compares canonicalized shapes of the same category with possibly different geometry -- note that PCA even outperforms ground truth canonicalization for this metric. Thus a method which is optimal for GC metric cannot outperform PCA in CC.

\parahead{PCA Under-Performance on the IC Metric}
The most likely reason why PCA underperforms on the IC metric is because of frame ambiguity.
The PCA principal directions are defined up to symmetries of the covariance matrix eigenspaces -- the shape does not necessarily share these symmetries.
For instance, when eigenvalues are distinct, eigenvectors are defined up to sign, causing random flips over principal directions: \eg,~an airplane can be flipped on its back.
When two or more eigenvalues are identical, eigenvectors are defined up to rotation, \eg,~in chairs, the major component can be from the left leg to the top right corner or bottom right leg to top left corner.
Thus, PCA canonicalization of rotated copies of a given shape may not be equal due to symmetries of the shape, resulting in higher Chamfer/IC error.


\section{Qualitative Results}
\begin{figure}[h]
\centering
  \includegraphics[width=0.9\textwidth]{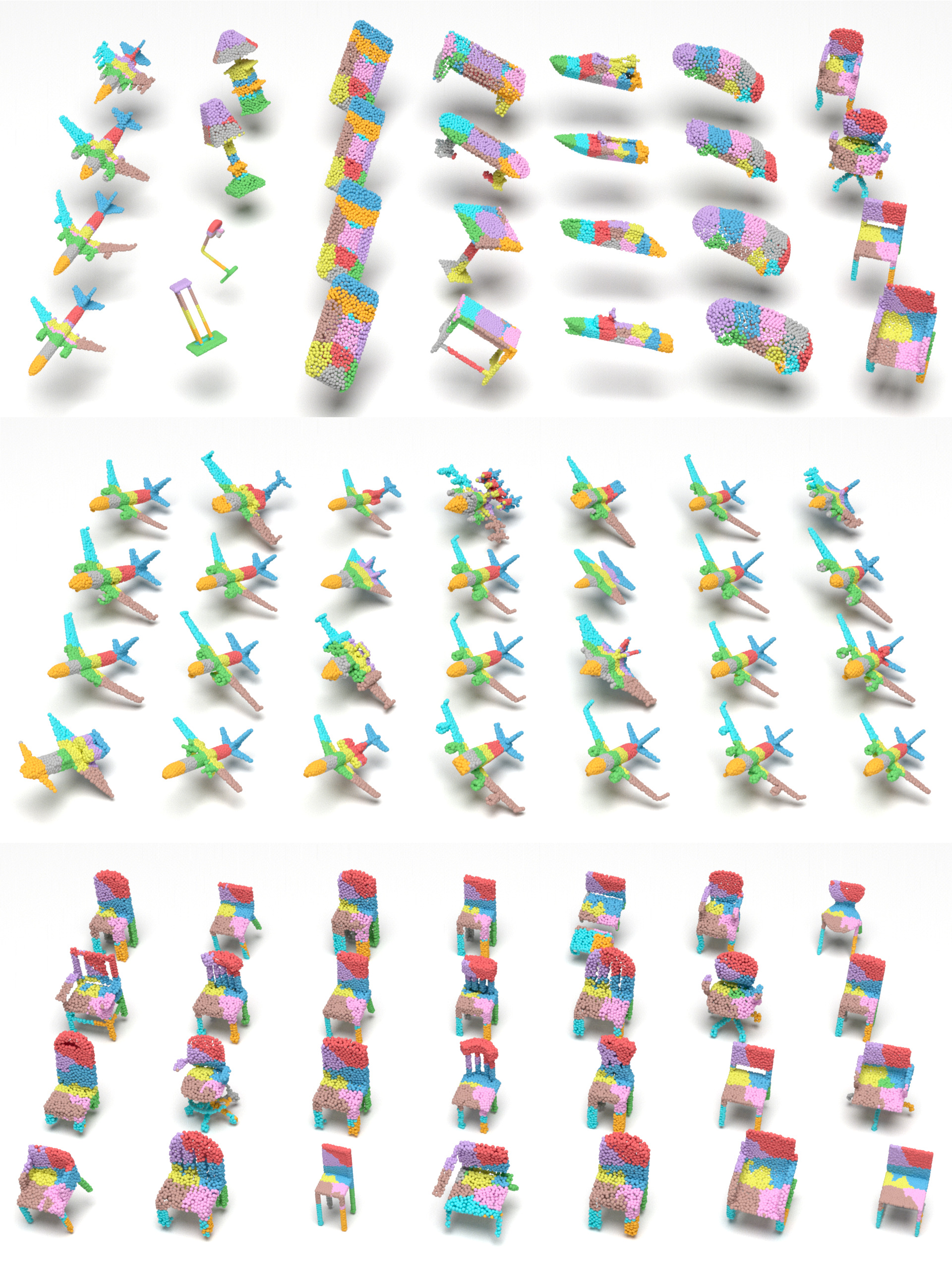}
  \caption{Parking lot for full shape canonicalization for multi-category(\textbf{\emph{top}}), plane (\textbf{\emph{middle}}) and chair (\textbf{\emph{bottom}}).  }
  \label{fig:qualitative_results_full_shapes}
\end{figure}

\begin{figure}[hbt]
\centering
\subfloat{\includegraphics[width=0.85\columnwidth]{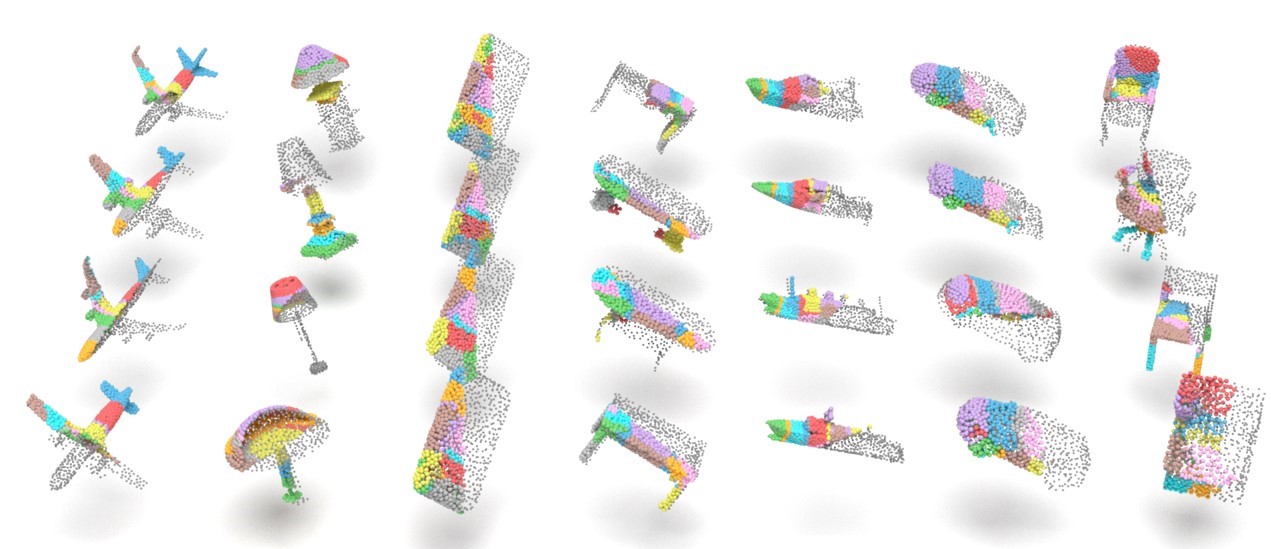}}
\vfill
\subfloat{\includegraphics[width=0.9\columnwidth]{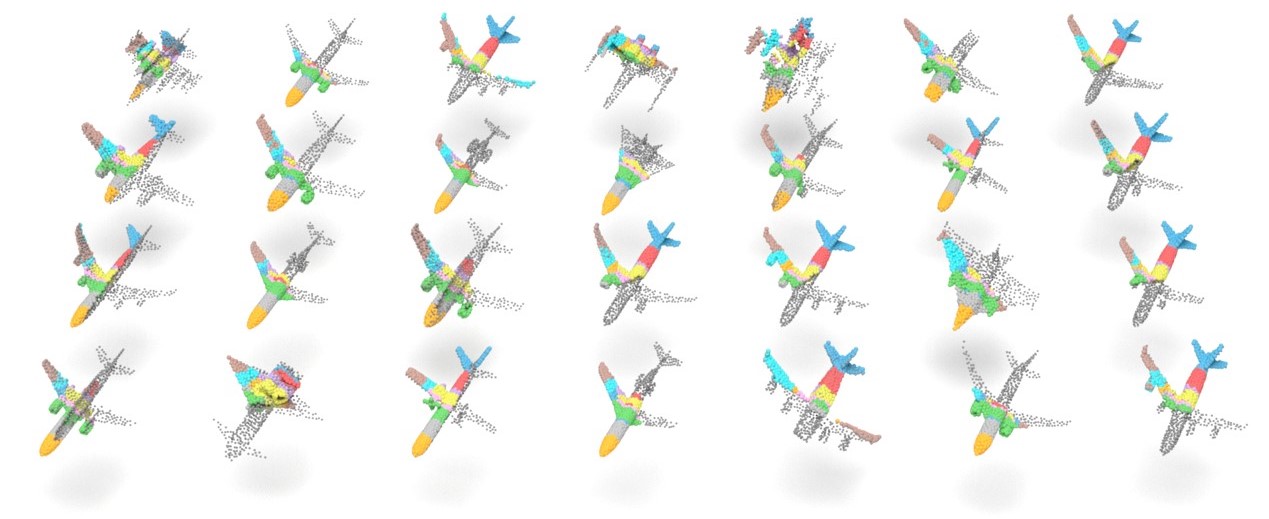}}
\vfill
\subfloat{\includegraphics[width=0.9\columnwidth]{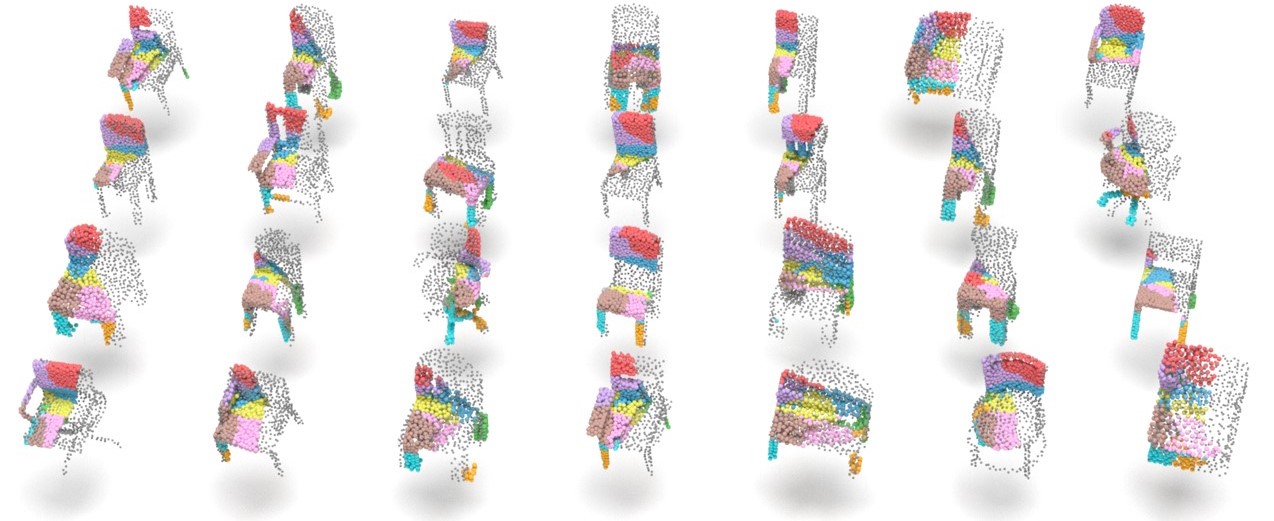}}
\caption{Parking lot for partial shape canonicalization for multi-category(\textbf{\emph{top}}), plane (\textbf{\emph{middle}}) and chair (\textbf{\emph{bottom}}). Note: missing parts only shown for visualization. }
\label{fig:qualitative_results_partial}
\end{figure}

We now present more qualitative results in Figure~
\ref{fig:qualitative_results_full_shapes}, ~\ref{fig:qualitative_results_partial} to demonstrate the effectiveness of our method. 
\onecolumn


\twocolumn

\end{document}